\documentclass{article}

\usepackage[preprint]{neurips_2026}

\usepackage{tikz}
\usetikzlibrary{arrows.meta, positioning, calc, shadows}
\usepackage[utf8]{inputenc}
\usepackage[T1]{fontenc}
\usepackage{hyperref}
\usepackage{url}
\usepackage{booktabs}
\usepackage{tabularx}
\usepackage{amsfonts}
\usepackage{amsmath,amssymb,amsthm}
\usepackage{nicefrac}
\usepackage{microtype}
\usepackage{xcolor}
\usepackage{graphicx}
\usepackage{mathtools}

\newcommand{\R}{\mathbb{R}}
\newcommand{\N}{\mathbb{N}}
\newcommand{\E}{\mathbb{E}}

\newcommand{\best}[1]{\textbf{#1}}

\newcommand{\secondbest}[1]{\underline{#1}}
\newcommand{\thirdbest}[1]{\textit{#1}}
\newcommand{\flagOOM}{\textsf{\footnotesize OOM}}

\title{xVGAE: A Hierarchical Variational Graph Autoencoder for Exchangeable Graphs}

\author{%
  Daniele Micheletti \\
  University of California, Irvine \\
  \texttt{dmichele@uci.edu} \\
  \And
  Federica Zoe Ricci \\
  Swarthmore College \\
  \texttt{fricci1@swarthmore.edu} \\
  \And
  Erik B.\ Sudderth \\
  University of California, Irvine \\
  \texttt{sudderth@uci.edu} \\
}
\begin{document}

\maketitle

\begin{abstract}
Graphs play a crucial role in many applications, ranging from drug discovery to social network modeling and astronomy. 
Generative models for graphs have substantially advanced in recent years, but we identify relatively simple scenarios where state-of-the-art models struggle to converge within a standard compute budget, or exhibit prohibitively slow generation times. 
We propose a generative architecture that achieves quality competitive with state-of-the-art models and is orders of magnitude faster at generation.
Our approach is inspired by the Aldous-Hoover theorem, a classic representation theorem that characterizes any probability distribution of graphs that is invariant to permutations of node indices or exchangeable. 
This theorem establishes that three ingredients suffice: a graph-level latent variable, a set of exchangeable node-specific latent variables, and a so-called  graphon function that maps graph- and node-level variables to the probability of an edge between any node pair.
Given a training set of graphs, our exchangeable Variational Graph AutoEncoder (xVGAE) learns an approximation to their underlying graphons, as well as graph- and node-level latent variables. We show that our xVGAE encodes latent representations that correlate with graph- and node-level properties, and can (quickly, in one shot) generate new graphs that closely match key training set statistics.
\end{abstract}

\section{Introduction}
\label{sec:intro}

Graphs provide mathematical descriptions of interconnected systems in fields ranging from biology \citep{boisset2018} to neuroscience \citep{bassett2017} and behavioral science \citep{aral2018}. While generative models for graphs offer substantial promise, their development is hindered by two primary challenges: (i) respecting the permutation invariance of node indices (assigning equal probability to isomorphic graphs), and (ii) balancing computational efficiency with the expressivity required to model potentially complex joint dependencies between all pairs of nodes.

Prior to recent advances in deep learning, statistical frameworks such as \emph{stochastic block models} (SBMs) \citep{holland1983, nowicki_estimation_2001}, \emph{mixed-membership} extensions (MMSBMs) \citep{airoldi2008}, \emph{random dot product graphs} (RDPGs) \citep{young2007}, and \emph{latent space models} (LSMs) \citep{hoff2002} were the primary generative models for networks.
These models structure the graph distribution via node-level and graph-level latent variables, a strategy that inherently addresses permutation invariance and maintains computational tractability. 
However, each of these models imposes restrictive assumptions on the structural graph characteristics captured by the latent variables, as well as the functional form governing edge probabilities.

A seminal effort to enhance the expressivity of these traditional generative models was the \emph{variational graph auto-encoder} (VGAE) \citep{kipf2016}, which employed \emph{graph convolutional networks} (GCNs) \citep{kipf2017} and inference methods for \emph{variational autoencoders} (VAEs) \citep{kingma2014} to generalize LSMs. While the original VGAE assumed an inflexible generative process and was primarily designed for representation learning, subsequent extensions introduced learnable decoders \citep{simonovsky2018, jin2018jtvae, samanta2020nevae}, though their experiments were limited to very small graphs and some training algorithms incorporated graph-matching procedures that are not scalable to larger graphs.

Among models capable of generating larger graphs, autoregressive architectures such as GraphRNN \citep{you2018graphrnn} and GRAN \citep{liao2019gran} construct adjacency matrices sequentially, sacrificing permutation invariance. More recently, diffusion-based approaches — both continuous \citep{niu2020, jo2022gdss} and discrete \citep{austin2021, hoogeboom2021multinomial, vignac2023, chen2023edge} — preserve permutation equivariance and have set new standards for sample quality by directly denoising graph structures. Despite rapid advances, the practical success of diffusion graph models remains largely confined to small graphs with rich metadata (e.g., molecules) or medium-sized but very sparse graphs (e.g., proteins, point clouds).

Current state-of-the-art models can become prohibitively slow during generation when applied to dense graphs of moderate size (hundreds of nodes). Methods prioritizing scalability often rely on sparsity assumptions that fail in denser settings \citep{bergmeister2024}, while highly expressive models suffer from generation bottlenecks because they either construct graphs sequentially \citep{you2018graphrnn, liao2019gran} or iteratively update explicit latent variables for every potential edge across many denoising steps \citep{niu2020, chen2023edge, vignac2023, zhou2024lgd}. Furthermore, the latent variables of existing diffusion and autoregressive models represent either noisy versions of the adjacency matrix or sequence-dependent connectivity profiles, failing to provide any semantic decomposition or interpretable compression of the graph.

In this work, we address these limitations by introducing the \emph{eXchangeable Variational Graph AutoEncoder} (xVGAE), a VAE-based generative model that adopts the Aldous-Hoover representation theorem for exchangeable graphs as its foundational design principle. This theoretical framework provides a principled way to decompose the structure of any exchangeable graph into node-level and graph-level latents without requiring per-edge latent variables. By adopting this design principle, our model achieves a dual benefit: it remains computationally tractable on dense graphs with hundreds of nodes—where existing methods often incur heavy training and sampling overhead—and it enables one-shot generation with quality comparable to competitors at speeds orders of magnitude faster. An added advantage of our approach is the ability to explore its learned latent representations, which we show correlate with key properties of community, ego, and protein networks.


\begin{figure}[t]
\centering
\resizebox{0.98\linewidth}{!}{%
    \begin{tikzpicture}[
    font=\sffamily,
    >=Stealth,
    encbg/.style={fill={rgb,255:red,116; green,130; blue,148}, draw={rgb,255:red,82; green,94; blue,110}, line width=1.1pt, rounded corners=6pt, drop shadow={opacity=0.18, shadow xshift=1pt, shadow yshift=-1.5pt}},
    encsub/.style={fill={rgb,255:red,219; green,233; blue,249}, draw={rgb,255:red,158; green,186; blue,222}, line width=0.9pt, rounded corners=3.5pt, minimum width=3.1cm, minimum height=1.15cm, align=center},
    decbg/.style={fill={rgb,255:red,184; green,219; blue,196}, draw={rgb,255:red,126; green,168; blue,140}, line width=1.1pt, rounded corners=6pt, drop shadow={opacity=0.18, shadow xshift=1pt, shadow yshift=-1.5pt}},
    decsub/.style={fill={rgb,255:red,248; green,252; blue,249}, draw={rgb,255:red,162; green,196; blue,172}, line width=0.9pt, rounded corners=4pt, align=center},
    optbox/.style={fill={rgb,255:red,232; green,240; blue,226}, draw={rgb,255:red,130; green,166; blue,138}, line width=0.9pt, dashed, rounded corners=4pt, minimum width=2.55cm, minimum height=0.92cm, align=center, font=\small},
    iobox/.style={fill={rgb,255:red,242; green,243; blue,245}, draw={rgb,255:red,90; green,98; blue,110}, line width=1.1pt, rounded corners=4pt, minimum width=1.25cm, minimum height=1.0cm, align=center, font=\large, drop shadow={opacity=0.15, shadow xshift=1pt, shadow yshift=-1pt}},
    latent/.style={circle, fill={rgb,255:red,246; green,194; blue,112}, draw={rgb,255:red,196; green,142; blue,64}, line width=1.1pt, minimum size=1.12cm, font=\large, drop shadow={opacity=0.15, shadow xshift=1pt, shadow yshift=-1pt}},
    plate/.style={draw={rgb,255:red,70; green,70; blue,70}, line width=0.75pt, rounded corners=3.5pt},
    infarrow/.style={->, dashed, draw={rgb,255:red,78; green,129; blue,189}, line width=1.25pt},
    priarrow/.style={->, draw={rgb,255:red,184; green,76; blue,62}, line width=1.25pt},
    decarrow/.style={->, draw={rgb,255:red,112; green,161; blue,122}, line width=1.25pt},
    priortext/.style={text={rgb,255:red,168; green,56; blue,43}, font=\normalsize}
]

\draw[encbg] (0, -2.35) rectangle (3.65, 2.35);
\node[font=\bfseries\large, text=black] at (1.825, 1.92) {Encoder};
\node[encsub] (enc_global) at (1.825, 0.82) {Global\\[2pt] $q_\phi(z_0 \mid A)$};
\node[encsub] (enc_local)  at (1.825, -1.08) {Local\\[2pt] $q_\phi(z_i \mid z_0, A)$};

\node[iobox] (A_in) at (-1.35, -0.13) {$A$};
\draw[infarrow] (A_in.east) -- (0, -0.13);
\draw[infarrow] (enc_global.south) -- (enc_local.north);

\node[latent] (z0) at (5.95, 0.98) {$z_0$};
\node[priortext, above=2pt of z0, xshift=14pt] {$p_\theta(z_0)$};

\node[latent] (zi) at (5.95, -1.08) {$z_i$};
\draw[plate] (5.12, -1.92) rectangle (6.78, -0.28);
\node[font=\small] at (6.48, -1.72) {$n$};

\draw[priarrow] (z0.south) -- (zi.north) node[midway, right=2pt, priortext] {$p_\theta(z_i \mid z_0)$};
\draw[infarrow] (enc_global.east) -- ++(0.65, 0) |- (z0.west);
\draw[infarrow] (enc_local.east) -- (zi.west);

\draw[decbg] (7.65, -2.35) rectangle (14.45, 2.35);
\node[font=\bfseries\large, text=black] at (11.05, 1.92) {Decoder};

\draw[decsub] (7.95, -1.98) rectangle (10.95, 1.15);
\node[font=\normalsize] at (9.45, 0.82) {Refiner};
\node[font=\large] at (9.45, 0.25) {$h_i$};
\node[font=\small] at (9.45, -0.25) {Node MLP + FiLM};
\node[optbox] (opt_att) at (9.45, -1.22) {Optional\\[-1pt] Attention};

\draw[decsub] (11.25, -1.98) rectangle (14.15, 1.15);
\node[font=\normalsize] at (12.70, 0.82) {Bilinear Head};
\node[font=\normalsize] at (12.70, -0.10) {$p_\theta(A_{ij} \mid h_i, h_j, z_0)$};
\node[font=\small] at (12.70, -0.78) {Edge Readout};

\draw[decarrow] (z0.east) -- ++(0.62, 0) -- ++(0, 0.62) -| (9.45, 1.15);
\draw[decarrow] (9.45, 1.60) -| (12.70, 1.15);
\draw[decarrow] (zi.east) -- (7.95, -1.08);
\draw[decarrow] (10.95, -0.25) -- (11.25, -0.25);

\node[iobox] (A_out) at (15.80, -0.13) {$\hat{A}$};
\draw[decarrow] (14.45, -0.13) -- (A_out.west);

\begin{scope}[yshift=-3.05cm]
    \draw[infarrow] (3.4, 0) -- (4.15, 0) node[right=3pt, text=black, font=\normalsize] {Inference};
    \draw[priarrow] (6.3, 0) -- (7.05, 0) node[right=3pt, text=black, font=\normalsize] {prior};
    \draw[decarrow] (8.55, 0) -- (9.30, 0) node[right=3pt, text=black, font=\normalsize] {decoder};
\end{scope}
\end{tikzpicture}}
\vspace*{-6pt}
\caption{\textbf{Overview of the xVGAE architecture (inference and generation).} The encoder (left) maps a graph $A$ to a global posterior $q_\phi(z_0 \mid A)$ and node-level posteriors $q_\phi(z_i \mid z_0, A)$ for the $n$ nodes. The prior (middle) samples the global latent $z_0 \sim p_\theta(z_0)$ and, conditionally, the independent node latents $z_i \sim p_\theta(z_i \mid z_0)$. The decoder (right) refines the node latents into features $h_i$ via an MLP, $z_0$-conditioned FiLM \citep{perez2018}, and optional attention. Finally, a bilinear head predicts pairwise edge probabilities $p_\theta(A_{ij} \mid h_i, h_j, z_0)$ for all node pairs to generate $\hat{A}$.}
\label{fig:architecture}
\end{figure}
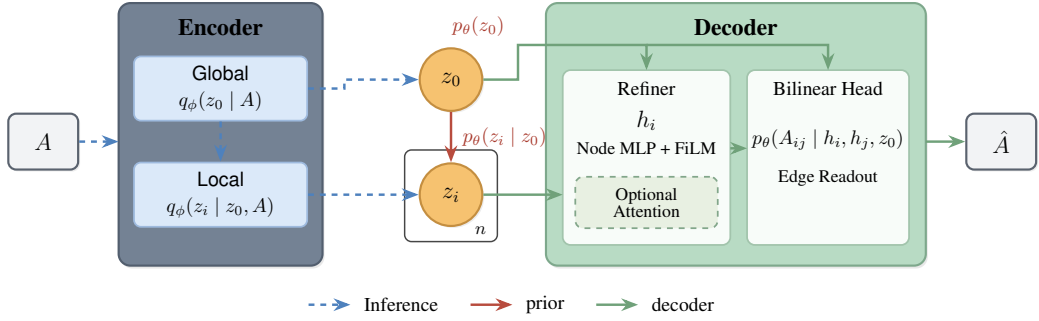

\section{Background: Representations of Exchangeable Graphs}
\label{sec:theory}

In this section, we review the theoretical foundations underlying the xVGAE architecture developed in Sec.~\ref{sec:method}. An undirected binary graph with $n$ nodes is represented by its adjacency matrix $A \in \{0,1\}^{n \times n}$. While graph sizes $n$ vary across the dataset, we denote the node count of any given graph simply by $n$ for clarity. We index nodes by $i, j \in \{1, \dots, n\}$, where $A_{ij} = A_{ji} = 1$ if an edge links nodes $i$ and $j$, and $0$ otherwise.

\paragraph{Permutation symmetry of graphs.}
The adjacency representation $A \in \{0,1\}^{n \times n}$ of a graph is not unique: permuting nodes' labels with a permutation $\pi$ results in a new adjacency matrix $A'$ such that $A'_{\pi(i), \pi(j)} = A_{ij}$, which corresponds to the same graph up to isomorphism. 
Any probabilistic model defined on adjacency matrices should respect this symmetry. 
Existing generative models of graphs have employed different approaches to account for it: permutation-equivariant graph neural networks build the symmetry into the architecture; equivariant diffusion models enforce it on each step of the noising and denoising processes. Autoregressive models are not invariant to permutations, but try to reduce artifacts by imposing (heuristically chosen) canonical node orderings, or by augmenting the training data with permutations of the graphs.

\paragraph{Joint exchangeability and the Aldous-Hoover theorem.} 
The probabilistic analogue of permutation invariance is \emph{joint exchangeability}. 
A random infinite 2d array $A = (A_{ij})_{i, j \in \N}$ is jointly exchangeable \citep{orbanz2014} if and only if, for every permutation $\pi$ of $\N$,
\begin{equation}
  (A_{ij})
  \stackrel{d}{=}
  (A_{\pi(i), \pi(j)}),
  \label{eq:joint-exch}
\end{equation}
where $\stackrel{d}{=}$ stands for equality in distribution, meaning that all permutations of the adjacency result in equal-probability graphs.
Aldous and Hoover \citep{aldous_representations_1981, hoover1979relations, kallenberg2005probabilistic} generalized the classic de Finetti theorem for exchangeable sequences to characterize jointly exchangeable 2d arrays. 
Such an array of a simple undirected graph is jointly exchangeable if and only if it admits the representation:
\begin{equation}
  A_{ij} \mid u_0, u_i, u_j \;\sim\; \mathrm{Bernoulli}\!\left(\omega(u_0, u_i, u_j)\right).
  \label{eq:graphon}
\end{equation}
Here, $u_0 \in \mathcal{U}$ is a \emph{global} or \textit{graph-level} latent variable; $u_i, u_j \in \mathcal{U}$ are \textit{local} or \emph{node-level} latent variables; and $\omega : \mathcal{U}^3 \to [0,1]$ is a measurable function called a \textit{graphon}, which is symmetric in $u_i$ and $u_j$. 
For each realization of $u_0$, the graphon $\omega(u_0, \cdot, \cdot)$ is deterministic, and the distribution of a set of exchangeable graphs is a mixture of graphons indexed by $u_0$.
In classic presentations of the Aldous-Hoover representation theorem, 
$u_0$ and $(u_i)_{i \in \N}$ are defined to be independent uniform random variables on $[0,1]$. 
However, any non-atomic distribution can take the place of the uniform, as $\omega$ can be adjusted via an appropriate inverse-CDF transform \citep{orbanz2014}. 
In Section \ref{sec:method}, we use this representation to construct a model that learns Euclidean (Gaussian or mixture) latent variables in place of uniformly-distributed ones, while maintaining exchangeability.

\paragraph{Connection to classical statistical models for graphs.} Classical latent variable models for graphs correspond to specific parametric restrictions for the graphon of Eq. \eqref{eq:graphon}: SBMs assign each node to one of finitely many communities and let edge probabilities depend on community memberships only; MMSBMs relax the hard assignment of SBMs by associating each node with a distribution over communities; RDPGs replace community labels with continuous node embeddings whose inner product determines the edge probability; LSMs \citep{hoff2002} extend this idea by passing the resulting score through a link function. In every case, the global variable $u_0$ controls the structure  shared across the graph ---  community interaction probabilities, distribution over embeddings  --- while the per-node variables $u_i$ encode each node's identity within that structure.

\paragraph{From representation to architecture.}
The Aldous--Hoover theorem is an existence result (for infinite exchangeable arrays) rather than a constructive prescription: it does not specify how to parameterize $\omega$, nor how to learn it from finite data. Yet, in a field with as many architectural degrees of freedom as deep generative modeling, a representation theorem provides a valuable structural inductive bias. It motivates organizing the latent space of a graph generative model around the two-level decomposition identified by the theorem: a global variable shared across the graph, and exchangeable per-node variables conditioned on it (see Fig.~\ref{fig:architecture}). A variational autoencoder that respects this hierarchy --- with a decoder playing the role of $\omega$ --- inherits a clean correspondence with the theory: the global latent indexes a graphon and the node latents index positions within it. Critically, the theorem indicates that this two-level latent structure is, in principle,  sufficient  to express rich exchangeable graph distributions without requiring explicit per-edge latent variables, provided the decoder is sufficiently flexible. We adopt this structure as the guiding design principle for our xVGAE model, whose concrete generative (decoder) and inference (encoder) architectures we develop in the next section.


\section{A Hierarchical VAE for Exchangeable Graphs}
\label{sec:method}

We now introduce xVGAE, a variational autoencoder whose latent structure is designed according to the Aldous--Hoover representation theorem for jointly exchangeable arrays (summarized in Fig.~\ref{fig:architecture}). In contrast to prior work drawing on this theorem for classical latent space models \citep{orbanz2014, lloyd2012random} or graphon estimation \citep{xia2023}, xVGAE adopts the Aldous--Hoover representation as its foundational framework for deep graph generation, empirically demonstrating its effectiveness on graphs with hundreds of nodes. 

\subsection{Generative model}
\label{sec:gen-model}
To take the roles of $u_0$ and $u_1, \dots , u_n$ from Eq. \eqref{eq:graphon}, we define learnable Euclidean latent variables: a graph-level vector $z_0 \in \R^{d_0}$ (the \textit{global} latent variable) and per-node vectors $z_1, \dots, z_n \in \R^{d}$ (the \textit{local} latent variables). A function $\omega_\theta : \R^{d_0} \times \R^{d} \times \R^{d} \to [0,1]$, symmetric in its last two arguments, determines the probability of each edge. Letting $\theta$ denote the set of all learnable parameters in the generative model,
\begin{equation}
  z_0 \sim p_{\theta}(z_0), \qquad
  z_i \mid z_0 \stackrel{\mathrm{iid}}{\sim} p_\theta(z_i \mid z_0), \qquad
  A_{ij} \mid z_0, z_i, z_j \sim \mathrm{Bernoulli}\!\left(\omega_\theta(z_0, z_i, z_j)\right).
  \label{eq:gen-model}
\end{equation}
Notice that the xVGAE local latent variables are independent and identically distributed conditionally on $z_0$, a parameterization allowed by the Aldous-Hoover theorem. Intuitively, this model formulates the graphon as a combination of two learnable parts: priors on per-graph and per-node latent variables that return Euclidean counterparts to the uniform latent variables of Eq. \eqref{eq:graphon}, and a decoder that maps the latent variables to edge probabilities. Next, we provide more details on these two parts.

\paragraph{Priors.}
For the conditional per-node prior distribution, we assume $p_\theta(z_i \mid z_0)$ is a Gaussian mixture with $K$ components whose weights, means, and diagonal covariances are determined by applying a deep neural network (in our experiments, a \emph{multi-layer perceptron} (MLP)) to $z_0$. The mixture form accommodates multimodal node distributions, such as community structure, while preserving exchangeability. Note that for $K=1$ we obtain a Normal distribution whose mean and variance are learnable functions of the global variable $z_0$. Marginalizing over $z_0$ yields a flexible prior on each $z_i$, so the global latent variable is responsible for mediating any dependence across nodes.
This structure reveals the pivotal role played by the global latent variable, which is also charged with guiding the decoder's mapping of per-node latents into edge probabilities (as detailed below). This motivates us to employ a flexible formulation for the graph-level prior $p_{\theta}(z_0)$, for which we adopt a latent diffusion process \citep{rombach2022high}. In practice, we find that a simple MLP score network and the DDPM framework \citep{ho2020} allow the model to learn a sufficiently expressive prior.
 
\paragraph{Decoder.}
The decoder parameterizes $\omega_\theta$ as a $z_0$-conditioned bilinear form on per-node embeddings,
\begin{equation}
  \omega_\theta(z_0, z_i, z_j) = \sigma\!\left( h_i^\top W(z_0)\, h_j + b(z_0,h_i,h_j)\right),
  \qquad h_i = \mathrm{MLP}\!\left(z_i;\, \mathrm{FiLM}(z_0)\right),
  \label{eq:graphon-head}
\end{equation}
where $\sigma$ is the logistic sigmoid activation function, $h_i \in \R^{d_h}$ is a per-node embedding produced by a shared MLP on $z_i$ whose layers are FiLM-modulated \citep{perez2018} by $z_0$, and $b(z_0,h_i,h_j) = b_0(z_0) + c(h_i) + c(h_j)$ are a global bias and a per-node bias $c$ shared by all nodes. $W(z_0) \in \R^{d_h \times d_h}$ is a symmetric interaction matrix parameterized by a MLP hypernetwork taking $z_0$ as input. The symmetry of $W(z_0)$ and $b(z_0,h_i,h_j)$ enforces the symmetry condition on $\omega_\theta$ required by the Aldous-Hoover Theorem (Sec.~\ref{sec:theory}), and the per-node application of the MLP keeps $h_i$ a function of $z_i$ and $z_0$ alone, so $\omega_\theta$ remains a pointwise graphon in the sense of Eq.~\eqref{eq:graphon}. More details on the technical parameterization of the bilinear decoder are discussed in Appendix \ref{sec:bilinear_decoder}. The classical models from Sec.~\ref{sec:theory} are recovered as special cases: a one-hot $h_i$ with $W(z_0)$ containing the log-odds of inter- and intra-community probabilities corresponds to a SBM, a simplex-valued $h_i$ to an MMSBM, and an identity $W$ to the inner-product decoder of the VGAE \citep{kipf2016}, a logistic version of the RDPG.

\paragraph{Optional extension: Set attention decoder.}
The form of Eq.~\eqref{eq:graphon-head} matches the graphon specified in Eq.~\eqref{eq:graphon}, up to the inverse-CDF identification of Euclidean $z_i$ with uniform variables noted above and exchangeable transformation. A natural extension that can increase the expressivity of the decoder is to replace the independent per-node MLPs with a series of 
\emph{self-attention blocks} (SAB, \citealt{lee2019}) operating on $\{z_1, \dots, z_n\}$ with FiLM conditioning on $z_0$, producing embeddings $\{h_i\}$ that may depend on the full set of node latents. SAB is permutation-equivariant, and applying a permutation-equivariant map to a conditionally i.i.d.\ collection of latent variables preserves their joint exchangeability, so the induced graph distribution remains exchangeable. 
This extension trades projectivity of the decoder for added flexibility. 

\subsection{Inference model}
VAEs are trained via an inference model or \emph{encoder}, which outputs an (amortized) variational approximation of the true posterior. We assume it factorizes to match the generative hierarchy,
\begin{equation}
  q_\phi(z_0, z_{1:n} \mid A) = q_\phi(z_0 \mid A)\, \prod_{i=1}^{n} q_\phi(z_i \mid A, z_0),
\end{equation}
with both factors (global and local encoder) parameterized as Normal distributions, whose means and (diagonal) covariances are produced by SABs acting on graph features. We use Laplacian eigenvectors and a small graph neural network to produce node features for our encoder input. The global encoder conditions the SAB layers on the normalized Laplacian eigenvalues through  FiLM layers, and pools its output through a permutation-invariant \emph{pooling by multihead attention} (PMA) module \citep{lee2019}. Similarly to the decoder, the local encoder is FiLM-conditioned on a sample of $z_0$ and outputs per-node Gaussian parameters. By construction, $q_\phi(z_0 \mid A)$ is permutation-invariant in the node ordering of $A$ and $q_\phi(z_{1:n} \mid A, z_0)$ is permutation-equivariant, mirroring the symmetries of the generative model. 

\subsection{Training}
Similar to many latent diffusion models~\citep{rombach2022high}, training occurs in two stages. First, we train the hierarchical VAE  to maximize the following two-level variational lower bound:
\begin{equation}
\begin{aligned}
  \mathcal{L}(\theta, \phi) \;=\;\;
  & \E_{q_\phi(z_0, z_{1:n} \mid A)}\!\left[\log p_\theta(A \mid z_0, z_{1:n})\right] \\
  & -\; \beta_0\,\mathrm{KL}\!\left(q_\phi(z_0 \mid A) \,\big\|\, p(z_0)\right) \\
  & -\; \beta\,\E_{q_\phi(z_0 \mid A)}\!\left[
    \mathrm{KL}\!\left(q_\phi(z_{1:n} \mid A, z_0) \,\big\|\, p_\theta(z_{1:n} \mid z_0)\right)
  \right].
\end{aligned}
\label{eq:elbo}
\end{equation}
In this first stage $p(z_0) = \mathcal{N}(0, I)$; the diffusion prior $p_\theta(z_0)$ is fitted afterwards. The first (reconstruction) term decomposes as a sum over the upper triangular entries of the symmetric adjacency matrix $A$, and the third (local KL) term decomposes as a sum over the $n$ nodes. Notice that training does not require sampling from the local prior, which would mean sampling the discrete assignments of the Gaussian mixture. This allows us to use a Monte Carlo approximation to get an unbiased estimate of the local KL term. The hyperparameters $\beta_0$ and $\beta$ may be tuned to improve the quality of the posterior approximation~\citep{higgins2017betavae}. In experiments, we typically fix \(\beta_0\leq1\), and initialize \(\beta\) to a small value that is gradually annealed towards 1. Appendix \ref{app:hyperparams} provides additional information on the annealing schedule used. 

Second, a small latent diffusion score network is trained using the encoded global embeddings \(z_0\) from the training data as target values, by minimizing the diffusion loss: \[
\mathcal{L}_{\mathrm{DM}} =
\mathbb{E}_{t, z_{0}^{(0)}, \epsilon}\!\Big[
\bigl\|\epsilon - \epsilon_\theta(z_{0}^{(t)}, t)\bigr\|_2^2
\Big],
\]
where \(\epsilon_\theta\) predicts the noise in \(z_{0}^{(t)}\). Similar to \cite{ho2020}, we optimize an unweighted diffusion loss rather than incorporating the weights required by the full ELBO~\citep{kingma2021variational}, which helps avoid training instabilities at low noise levels. Note that  latent diffusion is not a computational bottleneck for our model; this process is fast because we only have a single low-dimensional global variable for each graph in the training set. 

\subsection{Generation of New Graphs}
\label{sec:meth-generation}
A new graph with $n$ nodes is easily sampled by following the generative
process of Eq. \eqref{eq:gen-model}: draw
$z_0 \sim p_\theta(z_0)$ from the latent diffusion process, draw
$z_1, \dots, z_n \stackrel{\mathrm{iid}}{\sim} p_\theta(z_i \mid z_0)$
from the $z_0$-conditioned Gaussian mixture, and pass global and local latents
through the decoder in one forward pass, to obtain edge probabilities
$\omega_\theta(z_0, z_i, z_j)$. Edges are then drawn independently,
$A_{ij} \mid z_0, z_i, z_j \sim \mathrm{Bernoulli}(\omega_\theta(z_0, z_i, z_j))$. The node count $n$ can be set freely or drawn from the
empirical distribution of node counts in the training set, as we do in our experiments. 

\subsection{Visualization of the Learned Graph and Node Embeddings}
\label{sec:latent}

\begin{figure}
    \centering
    \includegraphics[width=0.98\linewidth]{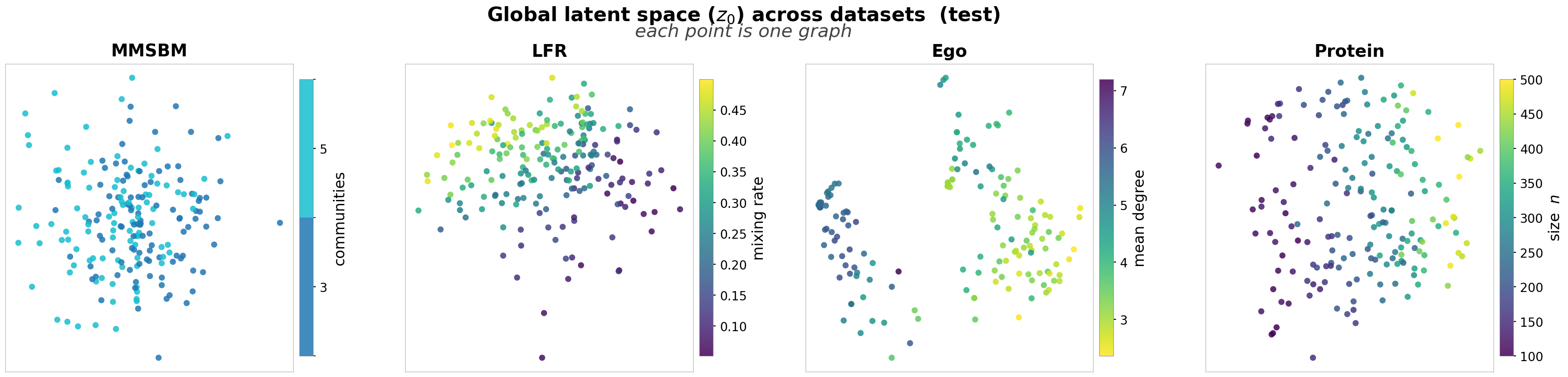}
    \caption{Two-dimensional PCA projections of the latent space for the global variable $z_0$ across all MMSBM, LFR, Ego, and Protein graphs in the test set. The latent space embeds structural information of the graphs: the plots show a clear association between the global variable and the graphs' number of communities, mixing rate $\mu$, average degree, and number of nodes $n$.}
    \label{fig:z0-stats}
\end{figure}

\begin{figure}
    \centering
    \includegraphics[width=0.98\linewidth]{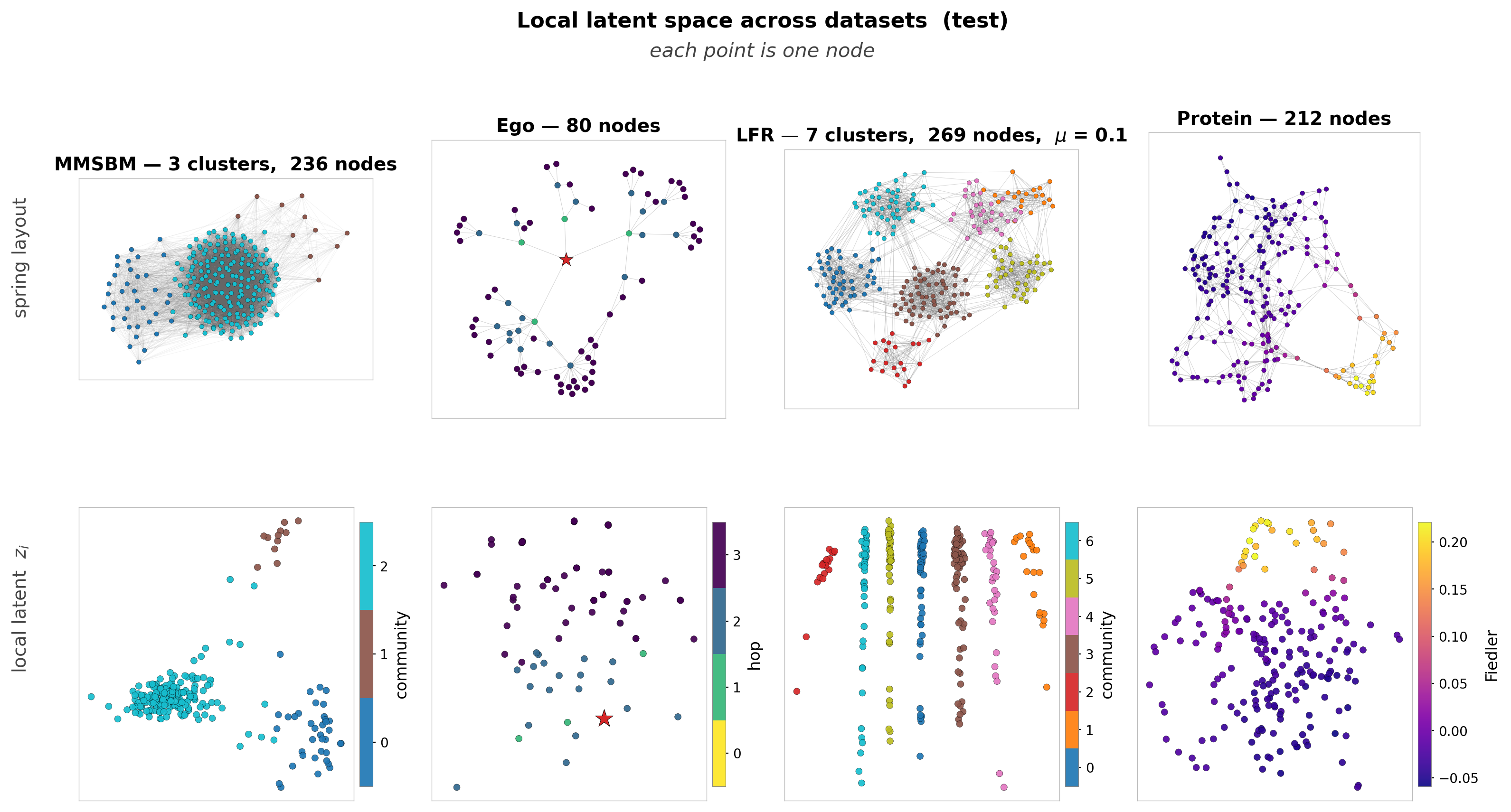}
    \caption{\textit{Top:} graphs from the test set of four different datasets (MMSBM, Ego, LFR, Protein) visualized using a spring layout. The nodes are colored by community membership for MMSBM and LFR, by distance from the ego node (represented by a star) for the Ego graph, and by Fiedler vector for Protein graphs. \textit{Bottom:} Two-dimensional PCA projections of the encoded local variables, which are associated with these node characteristics.}
    \label{fig:z-local}
\end{figure}

The two-level hierarchical xVGAE generative process incentivizes the latent variables to represent different graph characteristics: $z_0$ summarizes graph-level properties shared across all nodes, while each $z_i$ encodes the role of node~$i$ given this global context.  Unlike diffusion-based generative models that operate directly on graphs, where intermediate states are noisy versions of the adjacency matrix and admit no direct interpretation, both levels of our latent variables are inspectable on a per-graph and per-node basis. Figure \ref{fig:z0-stats} shows the correlation between graph statistics and the global latent space for four datasets. Figure \ref{fig:z-local} shows that the local latent variables capture node-level details such as group membership for MMSBM networks, and hop distance for Ego networks.

\section{Related Work}

\paragraph{Implicit Graphons.}
\cite{xia2023} parameterize graphons as coordinate-based MLPs to build an algorithm for graphon estimation, which recovers a graphon on uniform variables as in Eq.~\eqref{eq:graphon} given an observed graph. They learn no explicit generative model, and Gromov--Wasserstein alignment makes training slow. \citet{azizpour2024graphon} scale this approach to larger graphs, for representation learning and graph augmentation. We do not align nodes to a canonical uniform space; we instead use the Aldous--Hoover representation as an architectural guide for a hierarchical VAE aimed at fast, one-shot generation of new graphs.

\paragraph{VAEs for graph generation.}
The VGAE \citep{kipf2016} assumed a fixed decoder and was thus used for representation learning, but it inspired VAE-based generative models, first for small molecules \citep{simonovsky2018, jin2018jtvae, samanta2020nevae} and later via diffusion over continuous graph latents \citep{xu2023geometric}. LGD \citep{zhou2024lgd} embeds graphs with a permutation-equivariant transformer and trains a latent diffusion as a second stage; its unconditional generation results are limited to small graphs, and without a structured prior it needs both per-node and per-edge latents, which is costly on our benchmarks (Sec.~\ref{sec:experiments}). xVGAE instead structures the latent space into a global $z_0$ and conditionally i.i.d.\ node latents $z_i$, recovering the inductive biases of classical statistical models within the Aldous--Hoover representation.

\section{Experiments}
\label{sec:experiments}

\begin{figure*}[t]
    \centering
    \includegraphics[width=\textwidth]{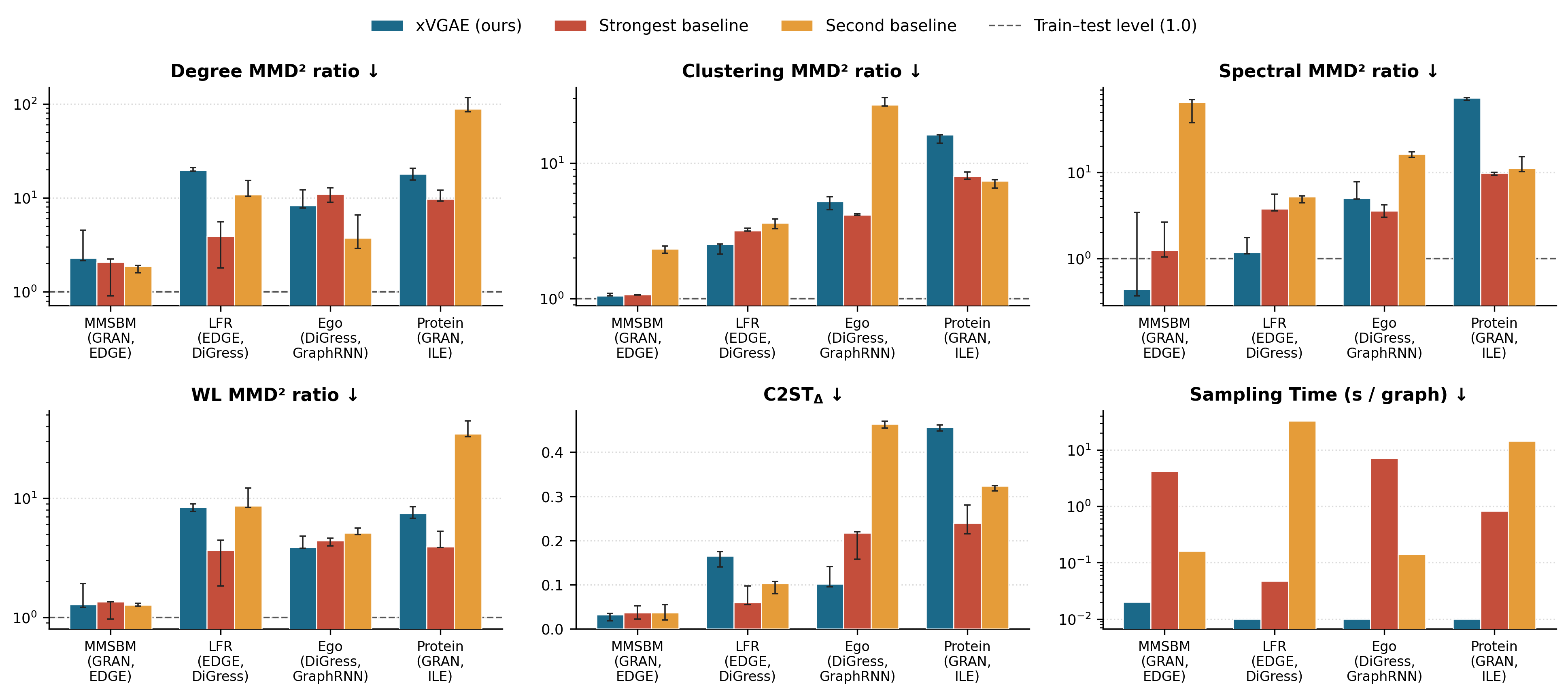}
    \vspace*{-12pt}
    \caption{\textbf{Benchmark overview across four datasets.} Panels compare \textbf{xVGAE} (blue) against the strongest (red) and second-strongest (orange) baselines per dataset. Both the xVGAE variant (MLP on MMSBM; SAB on LFR, Ego, and Protein) and the two baselines are selected per dataset by total column wins across degree, clustering, spectral, and WL $\mathrm{MMD}^2$ ratios (baseline names shown under each dataset). $\mathrm{MMD}^2$ ratios and $\mathrm{s/G}$ use a log scale; dashed lines mark the train--test reference ($1.0$). Lower is better; whiskers show min--max ranges over three draws.}
    \label{fig:overall_benchmarks}
\end{figure*}

We compare xVGAE against twelve graph generative baselines across four datasets spanning synthetic mixed-membership communities, power-law community networks, citation ego networks, and protein contact maps. We evaluate three variants of our model (hyperparameters and chunked bilinear decoder details in Appendix~\ref{app:hyperparams}): \textbf{xVGAE-MLP}, which uses the pointwise graphon head of Eq.~\eqref{eq:graphon-head}; \textbf{xVGAE-SAB}, which precedes the bilinear head with permutation-equivariant set-attention refinement (Sec.~\ref{sec:gen-model}); and \textbf{xVGAE without diffusion}, an ablation using a standard normal 
prior on $z_0$.

\subsection{Setup}
\label{sec:exp-setup}
\begin{table}[t]
\small
\caption{MMSBM ($n \in [120, 400]$, density $\sim\!27.5\%$). }
\centering
\label{tab:mmsbm-results}
\resizebox{\linewidth}{!}{%
\begin{tabular}{@{}l rrrr | rrr@{}}
\toprule
& \multicolumn{4}{c|}{$\mathrm{MMD}^2$ / train--test $\downarrow$} & \multicolumn{3}{c}{Auxiliary} \\
Model & deg & clu & spec & orb & WL $\downarrow$ & C2ST$_\Delta \downarrow$ & s/G $\downarrow$ \\
\midrule
\emph{Train vs.\ test (raw)} & 0.00179 & 0.00990 & 0.00016 & 0.01000 & 0.00714 & n/a & n/a \\
\midrule
\multicolumn{8}{@{}l}{\emph{One-shot}} \\
\quad VGAE & 85.7{\tiny\,[85.1--88.4]} & 31.2{\tiny\,[29.6--31.6]} & 1{,}002{\tiny\,[984--1{,}032]} & 1.00$^\dagger${\tiny\,[1.00--1.00]} & 25.4{\tiny\,[25.3--26.4]} & 0.470{\tiny\,[0.460--0.475]} & $<$0.01 \\
\quad GraphVAE$^*$ & 300 & 102 & 4{,}847 & 100$^\dagger$ & 166 & 0.500 & $<$0.01 \\
\quad SPECTRE & 9.19{\tiny\,[8.38--9.39]} & 2.81{\tiny\,[2.76--2.87]} & 6.00{\tiny\,[5.61--6.63]} & 1.00$^\dagger${\tiny\,[1.00--1.00]} & 2.28{\tiny\,[2.05--2.47]} & 0.195{\tiny\,[0.180--0.228]} & 0.03 \\
\quad IGNR & 22.2{\tiny\,[19.2--23.8]} & 21.3{\tiny\,[19.8--22.1]} & 14.0{\tiny\,[14.0--15.1]} & 1.00$^\dagger${\tiny\,[1.00--1.00]} & 7.75{\tiny\,[6.64--8.45]} & 0.238{\tiny\,[0.222--0.255]} & $<$0.01 \\
\addlinespace[2pt]
\multicolumn{8}{@{}l}{\emph{Autoregressive}} \\
\quad GraphRNN & 33.5{\tiny\,[32.0--34.3]} & 1.40{\tiny\,[1.39--1.48]} & 157{\tiny\,[157--168]} & 1.00$^\dagger${\tiny\,[1.00--1.00]} & 16.8{\tiny\,[16.0--16.8]} & 0.347{\tiny\,[0.320--0.350]} & 1.25  \\
\quad GRAN & \secondbest{2.07}{\tiny\,[0.91--2.24]} & 1.07{\tiny\,[1.06--1.07]} & \thirdbest{1.24}{\tiny\,[1.04--2.63]} & 1.00$^\dagger${\tiny\,[1.00--1.00]} & 1.35{\tiny\,[0.97--1.35]} & \secondbest{0.037}{\tiny\,[0.022--0.052]} & 4.15  \\
\addlinespace[2pt]
\multicolumn{8}{@{}l}{\emph{Diffusion}} \\
\quad EDP-GNN & 83.8{\tiny\,[23.0--90.0]} & 6.93{\tiny\,[2.60--8.24]} & 319{\tiny\,[16.8--356]} & 1.00$^\dagger${\tiny\,[1.00--1.00]} & 26.0{\tiny\,[9.93--28.1]} & 0.463{\tiny\,[0.120--0.463]} & 4.39  \\
\quad LGD$^*$ & 300 & 102 & 4{,}847 & 100$^\dagger$ & 166 & 0.500 & 26.6  \\
\quad DiGress & 74.3{\tiny\,[71.3--75.4]} & 2.78{\tiny\,[2.56--3.92]} & 846{\tiny\,[818--849]} & 1.00$^\dagger${\tiny\,[1.00--2.06]} & 26.4{\tiny\,[25.6--27.8]} & 0.356{\tiny\,[0.350--0.372]} & 18.2  \\
\quad SparseDiff & 24.1{\tiny\,[20.1--26.8]} & 3.49{\tiny\,[3.28--3.77]} & 186{\tiny\,[182--207]} & 1.00$^\dagger${\tiny\,[1.00--1.00]} & 11.3{\tiny\,[10.1--13.1]} & 0.218{\tiny\,[0.195--0.225]} & 45.2  \\
\quad EDGE & \best{1.86}{\tiny\,[1.60--1.90]} & 2.32{\tiny\,[2.16--2.44]} & 64.3{\tiny\,[37.8--69.3]} & 1.00$^\dagger${\tiny\,[1.00--1.00]} & \best{1.27}{\tiny\,[1.24--1.31]} & \thirdbest{0.037}{\tiny\,[0.020--0.055]} & 0.16  \\
\quad ILE & \multicolumn{7}{c}{\flagOOM}  \\
\midrule
\textbf{xVGAE-SAB} & 2.53{\tiny\,[2.48--4.31]} & \best{1.01}{\tiny\,[0.99--1.07]} & 1.86{\tiny\,[1.14--5.03]} & 1.00$^\dagger${\tiny\,[1.00--1.00]} & \thirdbest{1.34}{\tiny\,[1.25--1.76]} & 0.062{\tiny\,[0.018--0.065]} & $<$0.02 \\
\textbf{xVGAE-MLP} & \thirdbest{2.27}{\tiny\,[2.15--4.52]} & \thirdbest{1.05}{\tiny\,[1.05--1.09]} & \best{0.44}{\tiny\,[0.37--3.41]} & 1.00$^\dagger${\tiny\,[1.00--1.00]} & \secondbest{1.28}{\tiny\,[1.21--1.93]} & \best{0.032}{\tiny\,[0.018--0.035]} & $<$0.02 \\
\textbf{xVGAE} without diffusion & 3.47{\tiny\,[1.56--4.69]} & \secondbest{1.03}{\tiny\,[1.00--1.03]} & \secondbest{0.96}{\tiny\,[0.37--1.98]} & 1.00$^\dagger${\tiny\,[1.00--1.00]} & 1.47{\tiny\,[0.96--1.87]} & 0.040{\tiny\,[0.017--0.050]} & $<$0.01 \\
\bottomrule
\end{tabular}}
\footnotesize $^\dagger$ The summed orbit kernel saturates on this corpus (Appendix~\ref{app:orbit-lfr}). $^\ast$ Collapsed: degenerate samples (Table~\ref{tab:collapsed}).

\end{table}
\paragraph{Datasets.}
We evaluate on two synthetic and two real-world benchmarks (full specifications in Appendix~\ref{app:datasets}): \textbf{MMSBM} ($1{,}000$ graphs from~\citet{airoldi2008}, $n\!\in\![120,400]$, mean density~$27.5\%$, $K\!\in\!\{3,5\}$); \textbf{LFR}~\citep{lancichinetti2008benchmark} ($1{,}000$ graphs, $n\!\in\![200,500]$, mean density~$3.7\%$, per-graph mixing $\mu$ and power-law degree/community-size tails); \textbf{CiteSeer-Ego}~\citep{you2018graphrnn, martinkus2022, vignac2023} ($757$ $3$-hop ego networks from~\citet{sen2008collective}, $n\!\in\![50,399]$, mean density~$3.6\%$); and \textbf{Protein/D\&D}~\citep{dobson_distinguishing_2003} ($918$ protein graphs with heterogeneous size $n\!\in\![100,500]$, mean density~$2.3\%$).

\paragraph{Baselines and compute protocol.}
We compare against one-shot models (\textbf{VGAE}~\citep{kipf2016}, \textbf{GraphVAE}~\citep{simonovsky2018}, \textbf{SPECTRE}~\citep{martinkus2022}, \textbf{IGNR}~\citep{xia2023}), autoregressive generators (\textbf{GraphRNN}~\citep{you2018graphrnn}, \textbf{GRAN}~\citep{liao2019gran}), continuous and discrete diffusion models (\textbf{EDP-GNN}~\citep{niu2020}, \textbf{LGD}~\citep{zhou2024lgd}, \textbf{DiGress}~\citep{vignac2023}, \textbf{SparseDiff}~\citep{qin2024sparsediff}, \textbf{EDGE}~\citep{chen2023edge}), and iterative local expansion (\textbf{ILE}~\citep{bergmeister2024}). All models share a uniform budget (one 48\,GB GPU, 12\,h training, 10\,h sampling) using published hyperparameters. Under this fixed budget, computationally heavier baselines do not always reach their full published epoch counts and are evaluated at their best validation checkpoint (see Appendix~\ref{app:compute} for training protocols and a verification against published results). Generation time ($\mathrm{s/G}$) is the fastest per-graph runtime across batch sizes $1$, $8$, and the largest batch up to $256$ that fits in GPU memory (Appendix~\ref{app:sampling}). Because xVGAE runs diffusion only on the low-dimensional global vector $z_0$---followed by parallel node and pairwise decoding with a small memory footprint---  it supports large batch sizes without memory bottlenecks.

\paragraph{Metrics.}
Following~\citet{you2018graphrnn, martinkus2022, vignac2023}, we compute biased Gaussian-TV $\mathrm{MMD}^2$~\citep{gretton2012} on degree ($\mathrm{deg}$), clustering coefficient ($\mathrm{clu}$), normalized Laplacian spectrum ($\mathrm{spec}$), and 4-node orbit counts ($\mathrm{orb}$), along with Weisfeiler--Leman ($\mathrm{WL}$) subtree kernel $\mathrm{MMD}^2$~\citep{shervashidze2011} (detailed in Appendix~\ref{app:metrics}). Each $\mathrm{MMD}^2$ is reported as a ratio to the empirical train--test $\mathrm{MMD}^2$; a ratio near $1.0$ (or slightly below due to finite-sample variance in the subsampled train--test denominator) indicates that generated graphs match the test distribution within sampling noise. Because the summed orbit kernel saturates on dense or high-degree graphs (returning $\approx\!1.0$ for any graph distribution), saturated orbit columns are marked with $^\dagger$ (Appendix~\ref{app:orbit-lfr}). We also report the classifier two-sample test~\citep{lopezpaz2017, thompson2022}, which trains a discriminator to distinguish generated from real graphs; we report its accuracy deviation from random guessing, $\mathrm{C2ST}_\Delta = |\mathrm{acc} - 0.5|$, where $0$ means the two sets are indistinguishable (best) and $0.5$ means perfect separation (worst). Finally, we report generation time in seconds per graph ($\mathrm{s/G}$) considering sampling graphs in batches. All metrics are scored on the largest connected component across three draws (median and range reported); collapsed rows ($^\ast$, Table~\ref{tab:collapsed}) show a single value.

\begin{table}[t]
\centering
\small
\caption{CiteSeer-Ego ($n \in [50, 399]$, density $\sim\!3.6\%$).}
\label{tab:ego-results}
\setlength{\tabcolsep}{4pt}
\resizebox{\linewidth}{!}{%
\begin{tabular}{@{}l rrrr | rrr@{}}
\toprule
& \multicolumn{4}{c|}{$\mathrm{MMD}^2$ / train--test $\downarrow$} & \multicolumn{3}{c}{Auxiliary} \\
Model & deg & clu & spec & orb & WL $\downarrow$ & C2ST$_\Delta \downarrow$ & s/G $\downarrow$ \\
\midrule
\emph{Train vs.\ test (raw)} & 0.00031 & 0.00993 & 0.00095 & 0.01299 & 0.00437 & n/a & n/a \\
\midrule
\multicolumn{8}{@{}l}{\emph{One-shot}} \\
\quad VGAE$^*$ & 1{,}390 & 16.9 & 382 & 4.94 & 160 & 0.500 & $<$0.01 \\
\quad GraphVAE$^*$ & 2{,}469 & 107 & 719 & 77.1 & 368 & 0.500 & $<$0.01 \\
\quad SPECTRE & 42.1{\tiny\,[39.0--43.9]} & 5.62{\tiny\,[5.37--5.72]} & 11.1{\tiny\,[10.3--11.9]} & 2.48{\tiny\,[2.33--2.53]} & 11.5{\tiny\,[11.4--12.0]} & 0.260{\tiny\,[0.259--0.276]} & $<$0.01 \\
\quad IGNR & 79.1{\tiny\,[74.2--101]} & 14.3{\tiny\,[14.1--14.5]} & 47.1{\tiny\,[46.9--50.9]} & 2.97{\tiny\,[2.58--3.04]} & 23.7{\tiny\,[22.0--27.1]} & 0.359{\tiny\,[0.355--0.372]} & $<$0.01 \\
\addlinespace[2pt]
\multicolumn{8}{@{}l}{\emph{Autoregressive}} \\
\quad GraphRNN & \best{3.74}{\tiny\,[2.90--6.60]} & 26.8{\tiny\,[26.3--30.4]} & 16.2{\tiny\,[14.8--17.2]} & 4.35{\tiny\,[4.19--4.59]} & 5.11{\tiny\,[4.94--5.60]} & 0.464{\tiny\,[0.454--0.470]} & 0.14  \\
\quad GRAN & 27.2{\tiny\,[25.0--31.0]} & 7.73{\tiny\,[6.79--8.40]} & 13.9{\tiny\,[9.57--14.2]} & 3.50{\tiny\,[2.10--4.60]} & 7.32{\tiny\,[6.80--7.88]} & 0.270{\tiny\,[0.253--0.273]} & 0.46  \\
\addlinespace[2pt]
\multicolumn{8}{@{}l}{\emph{Diffusion}} \\
\quad EDP-GNN$^*$ & 512 & 12.5 & 88.6 & 7.08 & 79.0 & 0.441 & 54.0  \\
\quad LGD & 18.3{\tiny\,[17.2--18.7]} & 16.3{\tiny\,[15.7--16.9]} & 10.3{\tiny\,[9.01--10.4]} & 2.21{\tiny\,[2.10--2.26]} & 6.89{\tiny\,[6.78--7.00]} & 0.230{\tiny\,[0.217--0.243]} & 17.6  \\
\quad DiGress & 10.9{\tiny\,[8.92--12.8]} & \best{4.13}{\tiny\,[4.12--4.23]} & \best{3.57}{\tiny\,[3.00--4.21]} & \best{1.55}{\tiny\,[1.54--2.10]} & 4.41{\tiny\,[3.99--4.63]} & \thirdbest{0.217}{\tiny\,[0.158--0.220]} & 7.05  \\
\quad SparseDiff & 55.7{\tiny\,[53.1--59.1]} & 6.36{\tiny\,[6.10--6.46]} & 20.8{\tiny\,[20.1--21.0]} & 2.93{\tiny\,[2.71--3.04]} & 8.99{\tiny\,[8.75--9.14]} & 0.312{\tiny\,[0.303--0.329]} & 15.0  \\
\quad EDGE & 18.0{\tiny\,[16.9--26.9]} & 6.78{\tiny\,[6.33--7.25]} & 14.5{\tiny\,[12.2--18.7]} & 2.30{\tiny\,[2.22--2.86]} & 7.08{\tiny\,[6.98--9.40]} & \secondbest{0.148}{\tiny\,[0.112--0.165]} & 0.011  \\
\quad ILE & 21.7{\tiny\,[20.1--24.2]} & 5.85{\tiny\,[5.33--5.88]} & \thirdbest{5.90}{\tiny\,[5.60--6.35]} & 3.11{\tiny\,[3.02--3.63]} & 8.98{\tiny\,[8.50--9.02]} & 0.299{\tiny\,[0.277--0.346]} & 55.7  \\
\midrule
\textbf{xVGAE-SAB} & 8.30{\tiny\,[7.79--12.2]} & \secondbest{5.20}{\tiny\,[4.53--5.64]} & \secondbest{4.97}{\tiny\,[4.87--7.72]} & 2.62{\tiny\,[2.41--2.63]} & \secondbest{3.85}{\tiny\,[3.81--4.80]} & \best{0.102}{\tiny\,[0.095--0.141]} & $<$0.01 \\
\textbf{xVGAE-MLP} & \thirdbest{7.01}{\tiny\,[4.73--8.15]} & 7.01{\tiny\,[6.44--7.47]} & 12.0{\tiny\,[11.1--13.4]} & \secondbest{1.92}{\tiny\,[1.77--2.09]} & \thirdbest{3.95}{\tiny\,[3.49--4.08]} & 0.326{\tiny\,[0.326--0.335]} & $<$0.01 \\
\textbf{xVGAE} without diffusion & \secondbest{5.91}{\tiny\,[5.79--8.59]} & \thirdbest{5.81}{\tiny\,[5.24--6.23]} & 12.9{\tiny\,[10.9--12.9]} & \thirdbest{2.06}{\tiny\,[1.71--2.31]} & \best{3.57}{\tiny\,[3.52--3.92]} & 0.368{\tiny\,[0.355--0.382]} & $<$0.01 \\
\bottomrule
\end{tabular}}
\\[2pt]\footnotesize $^\ast$ Collapsed: degenerate samples (Table~\ref{tab:collapsed}).

\end{table}
\subsection{Results}
\label{sec:exp-results}

Figure~\ref{fig:overall_benchmarks} summarizes performance against the two top-ranked baselines per dataset, and Tables~\ref{tab:mmsbm-results}--\ref{tab:protein-results} report full results (best in \best{bold}, second \secondbest{underlined}, third \thirdbest{italicized}; \flagOOM\ denotes out-of-memory at batch size 1; $^\ast$ marks collapsed models).

\paragraph{MMSBM.}
On MMSBM (Table~\ref{tab:mmsbm-results}), xVGAE achieves the best clustering (SAB, $1.01$), spectral (MLP, $0.44$), and $\mathrm{C2ST}_\Delta$ (MLP, $0.032$) scores, and places a close second on WL (MLP, $1.28$ vs.\ $1.27$ for EDGE). Ratios near $1.0$ on spectrum and clustering indicate that xVGAE's generated graphs are close in distribution to the test graphs on these summaries, with $\mathrm{C2ST}_\Delta$ ($0.032$) similarly close to chance. Interestingly, xVGAE-MLP outperforms xVGAE-SAB on four of the five ranked metrics, suggesting that our pointwise Aldous--Hoover bilinear head is already well suited to dense mixed-membership community structures. While GRAN and EDGE also achieve strong quality, ILE runs out of GPU memory (\flagOOM) due to dense edge-triple message passing (Appendix~\ref{app:compute}), and several diffusion models (DiGress, SparseDiff, EDP-GNN, LGD) underperform within the 12\,h training budget while requiring up to $45$\,s/G to sample. Exploiting parallel batched decoding, xVGAE samples in $<\!0.02$\,s/G---roughly $8\times$ faster than EDGE and over $200\times$ faster than GRAN.

\paragraph{LFR.}
On LFR (Table~\ref{tab:lfr-results}), EDGE and DiGress are the strongest baselines overall, while ILE again exhausts GPU memory (\flagOOM). xVGAE-SAB achieves the best clustering ($2.50$) and spectral ($1.17$) $\mathrm{MMD}^2$ ratios overall, outperforming both EDGE and DiGress, though it trails them on degree, WL, and $\mathrm{C2ST}_\Delta$ (Fig.~\ref{fig:overall_benchmarks}). Unlike on MMSBM, xVGAE-SAB improves noticeably over xVGAE-MLP on clustering, spectrum, and WL, which suggests that set-attention refinement may help to model power-law degree and community-size variations. At generation time, xVGAE ($<\!0.01$\,s/G) is roughly $5\times$ faster than EDGE ($0.047$\,s/G), $100\times$ faster than GRAN ($1.03$\,s/G), and over $3{,}000\times$ faster than DiGress ($33.0$\,s/G).

\paragraph{CiteSeer-Ego.}
On CiteSeer-Ego (Table~\ref{tab:ego-results}), DiGress and GraphRNN are the two strongest baselines overall. Here, xVGAE-SAB achieves the best $\mathrm{C2ST}_\Delta$ ($0.102$ vs.\ $0.148$ for EDGE and $0.217$ for DiGress) and places second on clustering ($5.20$), spectrum ($4.97$), and WL ($3.85$, with our non-diffusion ablation reaching $3.57$), while outperforming DiGress on degree ($5.91$--$8.30$ vs.\ $10.9$; Fig.~\ref{fig:overall_benchmarks}). Furthermore, xVGAE generates graphs in $<\!0.01$\,s/G---nearly three orders of magnitude faster than DiGress ($7.05$\,s/G) and over $5{,}000\times$ faster than ILE ($55.7$\,s/G).

\paragraph{Protein.}
On Protein/D\&D (Table~\ref{tab:protein-results}), GRAN and ILE achieve the strongest overall sample quality, with SparseDiff winning on orbits. While xVGAE-SAB places second overall on degree ($17.9$) and WL ($7.46$) behind GRAN, it trails GRAN and ILE on clustering, spectrum, and $\mathrm{C2ST}_\Delta$ (Fig.~\ref{fig:overall_benchmarks}). Other latent-space and one-shot methods (LGD, IGNR, SPECTRE) similarly struggle on D\&D, suggesting that compressing highly irregular, chain-like protein geometries into fixed-dimensional global and local vectors limits fine-grained structural fidelity. Even so, xVGAE remains one to three orders of magnitude faster at sampling ($<\!0.01$\,s/G vs.\ $0.82$\,s/G for GRAN and $14.5$\,s/G for ILE; qualitative comparisons in Appendix~\ref{app:topologies}).

\paragraph{Meaningful latent variables.}
Figures~\ref{fig:z0-stats} and~\ref{fig:z-local} suggest that xVGAE's two-level hierarchy captures interpretable graph- and node-level structure without auxiliary supervision. The global latent $z_0$ correlates with key graph properties: community count $K$ on MMSBM, mixing $\mu$ on LFR, average degree on CiteSeer-Ego, and node count $n$ on Protein. Simultaneously, local latents $z_i$ reflect community membership on MMSBM and LFR, hop distance from the ego node on CiteSeer-Ego, and the Fiedler vector on Protein. See Appendix~\ref{app:latent} for extended visualizations.

\section{Conclusion}
\label{sec:conclusion}

We introduced an exchangeable variational graph autoencoder (xVGAE) whose latent structure is guided by the Aldous--Hoover representation theorem for jointly exchangeable arrays. To our knowledge, xVGAE is the first deep generative model built on the Aldous--Hoover representation to reach sample quality competitive with state-of-the-art baselines on graphs with hundreds of nodes, while sampling up to three orders of magnitude faster. The two-level latent decomposition (a graph-level $z_0$, per-node $z_i$ conditionally i.i.d.\ given $z_0$, and edges produced by a graphon-like map) yields a compact latent representation, with $z_0$ and $z_i$ empirically capturing global graph structure and node-level variation, respectively. This latent representation also allows rich generative processes, including denoising diffusions and mixture models, to be efficiently applied to graphs with hundreds of nodes. While the underlying representation theorem suggests that an xVGAE-style architecture can express rich exchangeable graph distributions, theoretical derivations are still needed to characterize the conditions under which a neural decoder can approximate arbitrary measurable symmetric kernels on finite graphs.

We evaluated our xVGAE on two complex synthetic benchmarks (MMSBM, LFR) and two real-world benchmarks (CiteSeer-Ego, Protein/D\&D) against twelve baselines spanning one-shot VAEs, GANs, graphon estimators, autoregressive models, continuous- and discrete-state diffusion, and iterative local expansion. xVGAE achieves sample quality competitive with or superior to the strongest baselines on MMSBM, LFR, and CiteSeer-Ego, and places second on degree and WL on Protein/D\&D, while generating graphs faster than every autoregressive and diffusion competitor on every dataset, from about $2\times$ (EDGE) to three orders of magnitude.

The Aldous--Hoover form admits extensions to separately exchangeable and partially exchangeable arrays \citep{orbanz2014}, opening a path for future generalizations of our xVGAE to bipartite, directed, and other asymmetric relational structures. While we focused strictly on binary adjacency matrices to isolate pure topological modeling, our framework can also be naturally extended to incorporate weighted edges, edge attributes, and node features or labels. On the architectural side, replacing the set-attention block with a linear-attention variant would drop the encoder and SAB-decoder cost from quadratic to linear in the number of nodes. 
However, even with the current asymptotically quadratic cost, the xVGAE's practical time and memory footprint is substantially lower than that of graph diffusion baselines.

Finally, while graph generative models offer powerful tools for simulating complex relational systems in fields such as network science and biology, their deployment requires careful consideration of the structural biases and distributional limitations inherent in training data, ensuring that generated synthetic graphs are used responsibly and transparently.

\newpage

\begin{ack}
This research was supported in part  by the HPI Research Center in Machine Learning and Data Science at UC Irvine.
\end{ack}

\bibliographystyle{plainnat}
\bibliography{references}

\appendix

\section{Datasets}
\label{app:datasets}

We evaluate on four graph-generation benchmarks spanning real-world
citation topology, synthetic community-structured graphs, synthetic
power-law community graphs, and biological protein-contact networks.
Table~\ref{tab:datasets-summary} summarises the key statistics;
the following subsections describe each dataset's origin, generation
process, split, and prior uses in the literature.

\begin{table}[h]
\centering
\caption{%
  Summary statistics for the four benchmarks.
  Splits are train / val / test graph counts.
  $n$, density $2|E|/(n(n-1))$ and mean degree $2|E|/n$ are per-graph averages over all splits;
  $\bar{E}$ is the mean edge count of the scored test graphs (largest connected component).
}
\label{tab:datasets-summary}
\resizebox{\textwidth}{!}{%
\begin{tabular}{lcccccc}
\toprule
\textbf{Dataset} & \textbf{Split (tr/va/te)} & $n$ \textbf{(min / mean / max)} & \textbf{Density} & \textbf{Mean degree} & $\bar{E}$ & \textbf{Type} \\
\midrule
CiteSeer-Ego   & 484 / 121 / 152 & 50 / 145 / 399  & 0.036 & 4.1  & 328  & Real, citation \\
MMSBM          & 640 / 160 / 200 & 120 / 249 / 400 & 0.275 & 68.4 & 9144 & Synthetic, blockmodel \\
LFR            & 640 / 160 / 200 & 200 / 350 / 500 & 0.037 & 12.2 & 2044 & Synthetic, power-law \\
Protein (D\&D) & 587 / 147 / 184 & 100 / 258 / 500 & 0.023 & 5.0  & 632  & Real, biological \\
\bottomrule
\end{tabular}%
}
\end{table}

\subsection{CiteSeer-Ego}
\label{sec:ds-ego}

\paragraph{Origin and generation.}
Three-hop ego-networks extracted from the CiteSeer citation graph
\citep{sen2008collective}.
For each seed node (paper), the induced subgraph is formed by
collecting all nodes within three BFS hops and retaining every edge
whose both endpoints lie within this neighbourhood.
The resulting graphs have a hub-and-spoke topology: the ego node is
linked to its immediate neighbours, which connect to sparser outer
layers.
Self-loops are removed and the graph is treated as undirected.

\paragraph{Node features.}
Each node is decorated with the leading $d{=}50$ eigenvectors of the
normalized graph Laplacian. A deterministic sign convention is applied (argmax-$|\cdot|$ column forced non-negative) for reproducibility across runs and machines. The eigenvalue vector $\lambda \in \mathbb{R}^d$ is stored alongside.

\paragraph{Split.}
484 / 121 / 152 (train / val / test), seed 42123.
Node-count filter: graphs with fewer than 50 or more than 399 nodes
are discarded before splitting.

\paragraph{Uses in the literature.}
Standard small-graph benchmark for unconditional graph generation.
Used by GraphRNN \citep{you2018graphrnn}, SPECTRE \citep{martinkus2022},
DiGress \citep{vignac2023}, GRAN \citep{liao2019gran}, and essentially
every modern graph generation paper as the canonical ``dense small
graphs'' benchmark.
Our version follows the GraphRNN preprocessing (3-hop, symmetric,
no self-loops) and the SPECTRE eigenvector-feature convention.

\subsection{MMSBM}
\label{sec:ds-mmsbm}

\paragraph{Origin and generation.}
Synthetic graphs sampled from a hierarchical instantiation of the
Mixed-Membership Stochastic Blockmodel (MMSB) of
\citet{airoldi2008}.
Each graph is generated independently as follows:
\begin{enumerate}
  \item Sample the community count $K$ uniformly from $\{3, 5\}$.

  \item Sample Dirichlet concentration
        $\alpha \sim \mathrm{Gamma}(2, 0.15)$ (per graph).
  \item For each node $i$, sample membership vector
        $\theta_i \sim \mathrm{Dir}(\alpha \cdot \mathbf{1}_K)$.
  \item Sample block affinities with mean
        $\bar{p}_{\mathrm{in}} \approx 0.5$ and
        $\bar{p}_{\mathrm{out}} \approx 0.025$,
        scaled by $\kappa{=}50$.
  \item For each node pair $(i,j)$: draw community indicators
        $z_i \sim \mathrm{Cat}(\theta_i)$,
        $z_j \sim \mathrm{Cat}(\theta_j)$,
        then $A_{ij} \sim \mathrm{Bern}(B_{z_i z_j})$ with
        $B_{kl} = p_{\mathrm{in}}$ if $k{=}l$, else $p_{\mathrm{out}}$.
  \item Sample graph size
        $n$ from a negative binomial with mean 250 and standard deviation 60, clipped to $[120, 400]$.
\end{enumerate}
Graphs are post-processed to retain only the largest connected
component (LCC).
Node features: leading $d{=}100$ normalised-Laplacian eigenvectors.

\paragraph{Split.}
1\,000 graphs total; 640 / 160 / 200, seed 4242.
Average density $\approx 27\%$, driven by the high
$p_{\mathrm{in}}$ and the moderate--large node counts.

\paragraph{Uses in the literature.}
Introduced in this work as a benchmark for graph generative models
with explicit per-graph community structure.
The MMSB generative process is well-studied in the community detection
literature \citep{airoldi2008}; using it as a generation benchmark is
novel to this paper.
The key challenge it poses is the per-graph variation in $K$ and in
membership concentration $\alpha$:
models without a graph-level latent variable collapse to a
dataset-mean community structure and cannot reproduce per-graph
variation in modular topology.
With $\bar{n}{=}249$ and density $0.275$ it is also the densest corpus
of the suite; ILE, whose state grows with the edges and triangles of
the graph, runs out of GPU memory on it
(Appendix~\ref{app:compute}).

\subsection{LFR}
\label{sec:ds-lfr}

\paragraph{Origin and generation.}
Lancichinetti--Fortunato--Radicchi (LFR) benchmark graphs
\citep{lancichinetti2008benchmark}, drawn with our own LFR sampler.
Each graph has power-law degree and community-size distributions and a
mixing parameter $\mu$, the fraction of each node's edges that leave
its community. All of these are drawn at random per graph:
\begin{itemize}
  \item $n \sim \mathrm{Uniform}\{200,\dots,500\}$;
  \item minimum degree $\sim \mathrm{Uniform}\{3,\dots,10\}$, maximum
        degree log-uniform on $[20,150]$ and at most $n/2$;
  \item degree exponent $\tau_1 \sim \mathrm{Uniform}[2,3]$ and
        community-size exponent $\tau_2 \sim \mathrm{Uniform}[1.1,2]$,
        community sizes in $[20, n/2]$;
  \item $\mu \sim \mathrm{Uniform}[0.05,0.5]$.
\end{itemize}
Degrees and community sizes are drawn exactly from these power laws,
stubs are matched inside and between communities, and
degree-preserving swaps make the graph simple. Every graph is checked
for its exact degree sequence, its internal degrees, its per-node mixing
and its connectivity; a graph that fails is redrawn.
Node features: leading $d{=}20$ normalised-Laplacian eigenvectors.

\paragraph{Split.}
640 / 160 / 200 (train / val / test). Training and validation graphs
are a seeded permutation of 800 graphs; the 200 test graphs are drawn
from the same generating process with seeds disjoint from training and
validation. Mean $n$ is 350 and mean density $0.037$
(Table~\ref{tab:datasets-summary}).

\paragraph{Uses in the literature.}
LFR is the standard community-detection benchmark
\citep{lancichinetti2008benchmark} used to evaluate algorithms such as
Louvain \citep{blondel2008}, Leiden \citep{traag2019}, Infomap, and
others; it is rarely used to evaluate generative models.
We include it because it combines heavy-tailed degrees, planted
communities, a per-graph mixing level and graphs of up to 500 nodes.

\subsection{Protein (D\&D)}
\label{sec:ds-protein}

\paragraph{Origin.}
The D\&D dataset of \citet{dobson_distinguishing_2003}.
Each graph represents a protein: nodes are amino acids, and edges
connect amino acids within a threshold Euclidean distance in the
folded three-dimensional structure.
The dataset is a standard graph classification benchmark, here repurposed as an \emph{unconditional generation} benchmark.

\paragraph{Generation and preprocessing.}
We use the version distributed with PyTorch Geometric
\citep{fey2019}.
Node labels (amino-acid type) are not used: like every model in our
tables, ours sees only the graph structure.
Graphs are filtered to the range $n \in [100, 500]$, yielding 918
graphs; the LCC is retained for all evaluation.
Node features: leading $d{=}100$ normalised-Laplacian eigenvectors.

\paragraph{Split.}
587 / 147 / 184 (train / val / test), seed 42123.

\paragraph{Uses in the literature.}
D\&D is used as a graph-classification benchmark in the Weisfeiler-Leman literature and in kernel-based methods. As a \emph{generation} benchmark it appears in GRAN \citep{liao2019gran}, SPECTRE \citep{martinkus2022} and ILE \citep{bergmeister2024}. It is the most heterogeneous dataset in our suite: node count ranges from 100 to 500 within a single split, and local density varies substantially between chain-dominated and core-dominated regions of the protein graph.

\subsection{Comparison: dataset structural regimes}

The four datasets cover four structurally distinct regimes:

\begin{itemize}
  \item \textbf{Small real-world.} CiteSeer-Ego: real topology,
    hub-and-spoke, high per-graph size variability, sparse.
    Widest baseline coverage in the literature.

  \item \textbf{Synthetic community structure.} MMSBM: planted
    mixed-membership block structure with known ground truth,
    moderate--large node count and high density.
    Tests per-graph $K$ and $\alpha$ variability under significant
    scale.

  \item \textbf{Power-law synthetic.} LFR: power-law degree and
    community-size tails, $n \in [200, 500]$, with the exponents and
    the mixing level varying from graph to graph.
    Stresses tail fidelity on the largest graphs of the suite on average.

  \item \textbf{Real-world heterogeneous.} Protein (D\&D): real
    biological graphs, large size range, heterogeneous density.
    Tests joint size-topology calibration under real-world
    structural variability.
\end{itemize}

\section{Metrics: Explanation and Discussion}
\label{app:metrics}

We evaluate generated graphs against the test set with six metrics.
Five are kernel-based two-sample tests (Maximum Mean Discrepancy,
$\mathrm{MMD}^2$); one is a classifier two-sample test ($\mathrm{C2ST}$).
All metrics are computed on the largest connected component (LCC)
of each graph.

\subsection{\texorpdfstring{$\mathrm{MMD}^2$}{MMD2} framework}

Given test set $\mathcal{X}$ and generated set $\mathcal{Y}$ with
graph-level feature maps $\phi(G) \in \mathbb{R}^d$, the biased
$\mathrm{MMD}^2$ estimator under kernel $k(\cdot, \cdot)$ is
\begin{equation}
  \widehat{\mathrm{MMD}}^2 = \frac{1}{|\mathcal{X}|^2}\!\!\sum_{x,x'\in\mathcal{X}}\!\!\!k(x,x') + \frac{1}{|\mathcal{Y}|^2}\!\!\sum_{y,y'\in\mathcal{Y}}\!\!\!k(y,y') - \frac{2}{|\mathcal{X}||\mathcal{Y}|}\!\!\sum_{x\in\mathcal{X},\,y\in\mathcal{Y}}\!\!\!k(x,y).
  \label{eq:mmd}
\end{equation}
We use the \emph{Gaussian total-variation} (TV) kernel
$k(x,y) = \exp\!\big(\!-\,\|x{-}y\|_{\mathrm{TV}}^2 / 2\sigma^2\big)$
with $\|x{-}y\|_{\mathrm{TV}} = \tfrac{1}{2}\sum_i |x_i {-} y_i|$,
following the GraphRNN/SPECTRE protocol \citep{you2018graphrnn, martinkus2022}.
The biased estimator is reported, to match the literature.

For the histogram-based metrics (degree, clustering and spectral), the
per-graph feature is a normalized histogram of a node-level statistic, on
a grid fixed before any sample is scored. The kernel keeps the published
bandwidths: $\sigma{=}1$ for degree and spectrum, $\sigma{=}0.1$ for clustering.

\paragraph{Ratios.} Every table reports $\widehat{\mathrm{MMD}}^2(\mathcal Y,\mathcal X_{\mathrm{test}})\,/\,\widehat{\mathrm{MMD}}^2(\mathcal X_{\mathrm{train}},\mathcal X_{\mathrm{test}})$, with the training set subsampled to the size of the test set and the denominator averaged over five subsamples; a ratio of 1 means the generated graphs are as close to the test graphs as held-out real graphs are \citep{martinkus2022}. We report the ratio for each statistic separately.

\subsection{Histogram metrics: degree, clustering, spectral}
\label{app:metrics-hist}

\paragraph{Degree MMD ($\mathrm{deg}$).}
Per graph: empirical histogram of node degrees, one bin per integer
degree.
Captures the marginal degree distribution.

\paragraph{Clustering MMD ($\mathrm{clu}$).}
Per graph: histogram of per-node local clustering coefficients
(triangle ratio over neighbour pairs), 100 bins on $[0,1]$.
Captures local triangle density.

\paragraph{Spectral MMD ($\mathrm{spec}$).}
Per graph: histogram of normalized graph-Laplacian eigenvalues,
100 bins on $[0,2]$.
Captures global mixing properties and modular structure
\citep{martinkus2022}.

\subsection{Orbit MMD}

\paragraph{Standard protocol (GraphRNN/SPECTRE).}
For each graph, count the number of times each node participates in
each of the 15 node orbits of graphlets with up to four nodes using ORCA
\citep{hovcevar2014combinatorial}, average across nodes to produce a single
15-dimensional graph-level feature vector, and apply
Eq.~\ref{eq:mmd} with a Gaussian-TV kernel ($\sigma{=}30$) on these
vectors. On MMSBM and LFR this kernel saturates (Appendix~\ref{app:orbit-lfr}). We report this standard protocol, so that the orbit column is comparable with prior work, and mark the corpora on which it saturates.

\subsection{\texorpdfstring{Weisfeiler--Leman MMD ($\mathrm{wl}$)}{Weisfeiler-Leman MMD (wl)}}

The graph-level feature is the Weisfeiler--Leman subtree feature
\citep{shervashidze2011}, computed as a hashed bag of WL labels after
four refinement rounds ($16{,}384$ buckets, $\ell_2$-normalised).
$\mathrm{MMD}^2$ is then computed in the WL feature space under a
linear kernel.
WL is the most stable ranker in our experience: it captures local
structural motifs at multiple scales and is robust to scale and
size differences across the four datasets.
We always include a WL column in every $\mathrm{MMD}^2$ table.

\subsection{\texorpdfstring{Classifier Two-Sample Test ($\mathrm{C2ST}$)}{Classifier Two-Sample Test (C2ST)}}

A logistic-regression binary classifier is trained to distinguish
real ($y{=}1$) from generated ($y{=}0$) graphs in the embedding of an
untrained three-layer GIN \citep{xu2019, thompson2022}, under 5-fold
cross-validation.
The reported score is the mean held-out accuracy:
$\mathrm{C2ST} = 0.5$ indicates real and generated graphs are
indistinguishable; $\mathrm{C2ST} = 1.0$ indicates the classifier
separates them perfectly.
Unlike $\mathrm{MMD}^2$, $\mathrm{C2ST}$ is a single scalar that
aggregates all features through a learned decision boundary. In tables we report the deviation from chance $\mathrm{C2ST}_\Delta = |\mathrm{acc} - \tfrac12|$, so that $\mathrm{C2ST}_\Delta = 0$ corresponds to indistinguishable distributions (best) and $\mathrm{C2ST}_\Delta = 0.5$ to perfect separability (worst).

\subsection{Reporting conventions}

\begin{itemize}
  \item All $\mathrm{MMD}^2$ values are reported in the biased form for
        consistency with the published baseline literature.
  \item All metrics are computed on the LCC of each graph;
  \item Lower is better for all five $\mathrm{MMD}^2$ columns and
        for $\mathrm{C2ST}_\Delta$ ($0$ = chance accuracy).
  \item Every $\mathrm{MMD}^2$ is reported as a ratio to the
        training-versus-test value of the same statistic.
  \item The orbit column uses the standard protocol; $^\dagger$ marks
        the corpora on which it saturates (Appendix~\ref{app:orbit-lfr}).
    \item A model is \emph{collapsed} on a dataset when its samples are degenerate: the largest connected component is empty or fragmented (mean size below a quarter of the test graphs'), or the graphs have at least ten times the test edge count (Table~\ref{tab:collapsed}). Collapsed rows are marked $^\ast$ and shown without ranges.

\end{itemize}


\begin{table}[t]
\centering
\small
\caption{LFR ($n \in [200, 500]$, density $\sim\!3.7\%$).}
\label{tab:lfr-results}
\setlength{\tabcolsep}{4pt}
\resizebox{\linewidth}{!}{%
\begin{tabular}{@{}l rrrr | rrr@{}}
\toprule
& \multicolumn{4}{c|}{$\mathrm{MMD}^2$ / train--test $\downarrow$} & \multicolumn{3}{c}{Auxiliary} \\
Model & deg & clu & spec & orb$^\dagger$ & WL $\downarrow$ & C2ST$_\Delta \downarrow$ & s/G $\downarrow$ \\
\midrule
\emph{Train vs.\ test (raw)} & 0.00075 & 0.00914 & 0.00013 & 0.00986 & 0.00405 & n/a & n/a \\
\midrule
\multicolumn{8}{@{}l}{\emph{One-shot}} \\
\quad VGAE$^*$ & 458 & 40.1 & 3{,}439 & 1.05 & 138 & 0.500 & $<$0.01 \\
\quad GraphVAE$^*$ & 862 & 110 & 5{,}744 & 102 & 351 & 0.500 & $<$0.01 \\
\quad SPECTRE$^\ast$ & 247 & 9.61 & 3{,}986 & 1.28 & 69.4 & 0.500 & 0.03 \\
\quad IGNR & 33.4{\tiny\,[32.7--36.8]} & 25.0{\tiny\,[24.0--26.0]} & 11.2{\tiny\,[11.0--11.8]} & 1.31{\tiny\,[1.29--1.35]} & 24.2{\tiny\,[23.9--27.0]} & 0.380{\tiny\,[0.360--0.380]} & $<$0.01 \\
\addlinespace[2pt]
\multicolumn{8}{@{}l}{\emph{Autoregressive}} \\
\quad GraphRNN & 70.1{\tiny\,[64.9--86.9]} & 12.4{\tiny\,[11.3--13.6]} & 9.78{\tiny\,[8.53--19.2]} & 1.16{\tiny\,[1.16--1.20]} & 40.8{\tiny\,[37.3--51.3]} & 0.465{\tiny\,[0.457--0.468]} & 0.18  \\
\quad GRAN & 20.7{\tiny\,[16.3--25.6]} & 3.69{\tiny\,[3.35--4.00]} & 8.49{\tiny\,[6.97--17.0]} & \thirdbest{1.04}{\tiny\,[1.04--1.05]} & 16.6{\tiny\,[13.9--18.8]} & \thirdbest{0.150}{\tiny\,[0.097--0.157]} & 1.03  \\
\addlinespace[2pt]
\multicolumn{8}{@{}l}{\emph{Diffusion}} \\
\quad EDP-GNN$^*$ & 423 & 19.8 & 3{,}475 & 1.15 & 121 & 0.500 & 68.2  \\
\quad LGD$^*$ & 308 & 13.3 & 3{,}074 & 9.86 & 103 & 0.490 & 37.5  \\
\quad DiGress & \secondbest{10.8}{\tiny\,[10.4--15.3]} & 3.61{\tiny\,[3.28--3.86]} & 5.22{\tiny\,[4.42--5.31]} & 1.04{\tiny\,[1.03--1.04]} & 8.64{\tiny\,[8.38--12.2]} & \secondbest{0.103}{\tiny\,[0.080--0.107]} & 33.0  \\
\quad SparseDiff & 31.3{\tiny\,[26.5--35.1]} & 3.60{\tiny\,[3.22--4.16]} & 86.4{\tiny\,[81.7--104]} & 1.05{\tiny\,[1.05--1.05]} & 21.3{\tiny\,[16.9--23.2]} & 0.325{\tiny\,[0.323--0.330]} & 65.3  \\
\quad EDGE & \best{3.88}{\tiny\,[1.79--5.55]} & \secondbest{3.16}{\tiny\,[3.14--3.30]} & 3.78{\tiny\,[3.57--5.55]} & 1.06{\tiny\,[1.06--1.08]} & \best{3.65}{\tiny\,[1.84--4.45]} & \best{0.060}{\tiny\,[0.055--0.097]} & 0.05  \\
\quad ILE & \multicolumn{7}{c}{\flagOOM}  \\
\midrule
\textbf{xVGAE-SAB} & 19.5{\tiny\,[19.2--21.0]} & \best{2.50}{\tiny\,[2.12--2.52]} & \best{1.17}{\tiny\,[1.13--1.74]} & 1.04{\tiny\,[1.00--1.06]} & \thirdbest{8.38}{\tiny\,[7.71--8.95]} & 0.165{\tiny\,[0.140--0.175]} & $<$0.01 \\
\textbf{xVGAE-MLP} & 22.0{\tiny\,[21.9--25.9]} & \thirdbest{3.55}{\tiny\,[3.13--3.82]} & \thirdbest{3.28}{\tiny\,[1.83--5.21]} & \best{1.02}{\tiny\,[1.00--1.04]} & 9.41{\tiny\,[8.55--10.5]} & 0.153{\tiny\,[0.140--0.155]} & $<$0.01 \\
\textbf{xVGAE} without diffusion &\thirdbest{16.8}{\tiny\,[15.8--16.8]}&5.39{\tiny\,[4.60--8.04]}&\secondbest{1.46}{\tiny\,[1.19--1.87]}&\secondbest{1.02}{\tiny\,[1.01--1.07]}&\secondbest{7.70}{\tiny\,[7.45--7.71]}&0.190{\tiny\,[0.162--0.200]}& $<$0.01 \\
\bottomrule
\end{tabular}}
\\[2pt]
\footnotesize $^\dagger$ The summed orbit kernel saturates on this corpus (Appendix~\ref{app:orbit-lfr}). \\\footnotesize $^\ast$ Collapsed: degenerate samples (Table~\ref{tab:collapsed}).

\end{table}

\begin{table}[t]
\centering
\small
\caption{Protein/D\&D ($n \in [100, 500]$, density heterogeneous).}
\label{tab:protein-results}
\setlength{\tabcolsep}{4pt}

\resizebox{\linewidth}{!}{%
\begin{tabular}{@{}l rrrr | rrr@{}}
\toprule
& \multicolumn{4}{c|}{$\mathrm{MMD}^2$ / train--test $\downarrow$} & \multicolumn{3}{c}{Auxiliary} \\
Model & deg & clu & spec & orb & WL $\downarrow$ & C2ST$_\Delta \downarrow$ & s/G $\downarrow$ \\
\midrule
\emph{Train vs.\ test (raw)} & 0.00022 & 0.01058 & 0.00031 & 0.00318 & 0.00174 & n/a & n/a \\
\midrule
\multicolumn{8}{@{}l}{\emph{One-shot}} \\
\quad VGAE & 2{,}541{\tiny\,[2{,}456--2{,}544]} & 28.7{\tiny\,[28.5--29.0]} & 1{,}012{\tiny\,[995--1{,}032]} & 238{\tiny\,[238--238]} & 623{\tiny\,[610--625]} & 0.500{\tiny\,[0.500--0.500]} & 0.006 \\
\quad GraphVAE & 2{,}956{\tiny\,[2{,}955--2{,}961]} & 100{\tiny\,[100--101]} & 536{\tiny\,[536--537]} & 444{\tiny\,[444--444]} & 970{\tiny\,[970--971]} & 0.500{\tiny\,[0.500--0.500]} & $<$0.01 \\
\quad SPECTRE$^*$ & 1{,}330 & 88.5 & 1{,}790 & 238 & 395 & 0.500 & 0.02 \\
\quad IGNR & 597{\tiny\,[562--597]} & 19.1{\tiny\,[18.5--19.2]} & 149{\tiny\,[148--151]} & 218{\tiny\,[209--218]} & 152{\tiny\,[146--153]} & 0.497{\tiny\,[0.497--0.497]} & $<$0.01 \\
\addlinespace[2pt]
\multicolumn{8}{@{}l}{\emph{Autoregressive}} \\
\quad GraphRNN & 29.2{\tiny\,[26.5--34.4]} & 11.1{\tiny\,[11.0--11.4]} & 54.4{\tiny\,[54.0--57.2]} & 135{\tiny\,[132--149]} & 13.5{\tiny\,[12.2--16.1]} & 0.440{\tiny\,[0.435--0.446]} & 0.12  \\
\quad GRAN & \best{9.67}{\tiny\,[9.21--12.0]} & \secondbest{7.93}{\tiny\,[7.57--8.56]} & \best{9.72}{\tiny\,[9.22--9.96]} & \thirdbest{64.1}{\tiny\,[58.3--76.1]} & \best{3.91}{\tiny\,[3.85--5.29]} & \best{0.239}{\tiny\,[0.215--0.280]} & 0.82  \\
\addlinespace[2pt]
\multicolumn{8}{@{}l}{\emph{Diffusion}} \\
\quad EDP-GNN & 177{\tiny\,[170--197]} & 49.2{\tiny\,[47.0--51.2]} & 207{\tiny\,[206--208]} & 332{\tiny\,[324--333]} & 55.8{\tiny\,[53.1--61.9]} & 0.489{\tiny\,[0.486--0.492]} & 68.2 \\
\quad LGD$^*$ & 1{,}741 & 102 & 2{,}391 & 239 & 485 & 0.500 & 30.5  \\
\quad DiGress & 196{\tiny\,[185--224]} & \thirdbest{8.22}{\tiny\,[8.14--8.39]} & \thirdbest{14.2}{\tiny\,[13.3--15.4]} & 200{\tiny\,[196--212]} & 60.8{\tiny\,[58.6--70.0]} & \thirdbest{0.391}{\tiny\,[0.389--0.399]} & 19.4  \\
\quad SparseDiff & 85.2{\tiny\,[77.8--91.7]} & 10.2{\tiny\,[9.37--11.1]} & 16.8{\tiny\,[14.6--18.9]} & \best{8.72}{\tiny\,[7.57--9.99]} & 36.5{\tiny\,[33.1--38.6]} & 0.418{\tiny\,[0.413--0.429]} & 36.8  \\
\quad EDGE & 91.9{\tiny\,[77.9--93.0]} & 18.9{\tiny\,[17.5--19.0]} & 112{\tiny\,[109--112]} & 350{\tiny\,[339--351]} & 31.5{\tiny\,[26.6--31.9]} & 0.435{\tiny\,[0.429--0.446]} & 0.022  \\
\quad ILE & 88.3{\tiny\,[82.2--117]} & \best{7.38}{\tiny\,[6.53--7.53]} & \secondbest{11.1}{\tiny\,[10.2--15.2]} & \secondbest{48.3}{\tiny\,[44.8--60.0]} & 34.8{\tiny\,[32.9--44.6]} & \secondbest{0.323}{\tiny\,[0.312--0.324]} & 14.5  \\
\midrule
\textbf{xVGAE-SAB} & \secondbest{17.9}{\tiny\,[15.5--20.5]} & 16.1{\tiny\,[14.0--16.1]} & 72.9{\tiny\,[68.4--73.7]} & 191{\tiny\,[173--212]} & \secondbest{7.46}{\tiny\,[6.77--8.46]} & 0.456{\tiny\,[0.448--0.462]} & $<$0.01 \\
\textbf{xVGAE-MLP} & 113{\tiny\,[111--115]} & 14.5{\tiny\,[13.8--14.9]} & 74.8{\tiny\,[74.3--75.0]} & 201{\tiny\,[192--202]} & 34.9{\tiny\,[33.7--35.7]} & 0.484{\tiny\,[0.478--0.484]} & $<$0.01 \\
\textbf{xVGAE} without diffusion & \thirdbest{19.4}{\tiny\,[18.1--20.2]} & 13.9{\tiny\,[13.5--15.5]} & 72.0{\tiny\,[70.4--72.5]} & 205{\tiny\,[174--210]} & \thirdbest{7.63}{\tiny\,[7.57--8.22]} & 0.451{\tiny\,[0.446--0.459]} & $<$0.01 \\
\bottomrule
\end{tabular}}
\\[2pt]
\footnotesize $^\ast$ Collapsed: degenerate samples (Table~\ref{tab:collapsed}).

\end{table}

\clearpage
\section{Experiments: extended details}
\label{app:experiments}
\subsection{Compute limits, protocol and baseline notes}
\label{app:compute}

\paragraph{Limits.} Every model, ours included, gets one GPU, 64\,GiB of host RAM, 12\,h of training and 10\,h of sampling, on a 48\,GB NVIDIA RTX 6000 Ada (some training ran on an A6000 of the same memory). Out of memory (\flagOOM) is reported only when measured at batch size 1 within these limits. This happens for ILE alone, on MMSBM and LFR: training fails with a CUDA out-of-memory error in the sparse PPGN message aggregation over edge triples, whose number is large on these dense graphs (MMSBM: 5.4\,GiB requested with 45.0\,GiB already allocated; LFR: 3.6\,GiB with 43.2\,GiB, after 54 minutes). DiGress had a bug that would put it OOM on larger datasets like MMSBM or LFR, more details below. 

\paragraph{Undertrained within the limits.} LGD on CiteSeer-Ego (373 of the authors' 3000 diffusion epochs), SPECTRE (about 23 epochs), SparseDiff (about 125 epochs) and EDGE (about 156 epochs) on LFR, and EDGE on MMSBM reach the 12\,h training limit before converging; they are reported at the checkpoint chosen on validation among those saved. GRAN completes the authors' $3000$ epochs on every dataset within the limits (MMSBM: about 7\,h; LFR: about 5\,h); its checkpoints are chosen on validation (MMSBM: epoch 2925; LFR: epoch 1725).

\paragraph{Fairness of the budget.} Some baselines are undertrained relative to the settings of their own papers. The comparison is nevertheless fair under our protocol: every model gets the same limits (one GPU, 12\,h of training, 10\,h of sampling), our model and most baselines converge well within them, and we prefer comparing more models to giving some of them more time. To check the implementations, we re-scored each baseline under the protocol of a published result (raw $\mathrm{MMD}^2$ of degree, clustering and orbits with the authors' evaluator, full graphs, one draw; Table~\ref{tab:published-check}). 7 out of 12 comparisons match; the mismatches are attributed mostly to the training budget (suspected on most of them by comparing the completed epochs to the ones suggested by the authors). One caveat: our CiteSeer-Ego split differs from the published ones.

\begin{table}[h]
\centering
\caption{Baselines against their published numbers: raw $\mathrm{MMD}^2$ degree / clustering / orbit under the published protocol (EMD kernel where marked, TV otherwise). Sources in parentheses: the table of the paper reporting the number.}
\label{tab:published-check}
\resizebox{\linewidth}{!}{%
\begin{tabular}{llllll}
\toprule
Baseline & Dataset & Published (source) & Ours, their protocol & Verdict & Likely cause \\
\midrule
GRAN & Protein & 2.0e-3 / 0.049 / 0.13 (GRAN T1) & 2.5e-3 / 0.082 / 0.16 & match & \\
GRAN & Ego & 7.7e-3 / 0.066 / 0.043 (HiGen T9) & 0.017 / 0.11 / 0.061 & match ($\le 2.2\times$) & split \\
GraphRNN & Protein & 4.0e-3 / 0.15 / 0.59 (SPECTRE T2) & 7.0e-3 / 0.12 / 0.40 & match & \\
GraphRNN & Ego (EMD) & 0.077 / 1.15 / 0.11 (EDGE T2) & 0.18 / 1.08 / 0.052 & match & \\
DiGress & Ego & 8.9e-3 / 0.054 / 0.030 (SparseDiff T1) & 3.9e-3 / 0.041 / 0.027 & match & \\
DiGress & Protein & 4.1e-3 / 0.049 / 0.13 (ILE T2) & 0.050 / 0.080 / 0.58 & mismatch, 4--12$\times$ & undertrained \\
EDGE & Ego (EMD) & 0.058 / 0.18 / 0.052 (EDGE T2) & 0.024 / 0.14 / 0.026 & match & \\
EDGE & Protein & 0.19 / 0.34 / 0.68 (ILE T2) & 0.023 / 0.20 / 1.08 & match & \\
SparseDiff & Protein & 1.5e-3 / 0.034 / 5.3e-3 (SparseDiff T7) & 0.018 / 0.10 / 0.025 & mismatch, 3--12$\times$ & undertrained \\
SparseDiff & Ego & 3.7e-3 / 0.032 / 0.020 (SparseDiff T7) & 0.018 / 0.065 / 0.039 & mismatch, 2--5$\times$ & undertrained; split \\
ILE & Protein & 3.0e-3 / 0.031 / 4.7e-3 (ILE T2) & 0.017 / 0.080 / 0.13 & mismatch, 3--28$\times$ & undertrained \\
SPECTRE & Protein & 5.6e-3 / 0.084 / 0.027 (SPECTRE T2) & 0.34 / 0.94 / 0.73 & mismatch, 30--60$\times$ & undertrained \\
\bottomrule
\end{tabular}}
\end{table}

\paragraph{Protocol.} Baselines run from their public code (VGAE and GraphVAE from our re-implementations) with upstream hyper-parameters, changed only where our graph sizes force it. All models share the scoring code, the splits and the selection rule: draws of one checkpoint are averaged, the checkpoint with the lowest validation score (geometric mean over the five statistics of $\mathrm{MMD}^2$ divided by its train-vs-validation value) wins, every candidate within $10\%$ of it is tied, and the tie goes to the C2ST accuracy closest to $0.5$; only the winner is scored on test. Each table row is the median of three sampling draws of that checkpoint, with their range. Collapsed rows (Appendix~\ref{app:metrics}) are drawn once. For xVGAE this rule picks both the autoencoder and the $z_0$ diffusion. The xVGAE model without diffusion is the best on validation among the MLP and SAB versions used for the model with diffusion and new MLP and SAB models trained with both $\beta$ and $\beta_0$ set to 1 for fairness. Every model draws $n$ from the training sizes, except GraphRNN and GraphVAE, which choose it themselves. VGAE, GraphVAE and LGD threshold edge probabilities at $0.5$; the others sample them. EDGE, as published, steers generation toward the degree sequence of a training graph, so it does not have to learn the degree distribution. EDP-GNN runs $6\times1000$ annealed Langevin steps ($6\times100$ on MMSBM). 

\paragraph{Collapsed models.} Table~\ref{tab:collapsed} lists the collapsed rows and the reason; all have $\mathrm{C2ST}$ accuracy close to $1$ ($\mathrm{C2ST}_\Delta \geq 0.44$).

\begin{table}[h]
\centering
\small
\caption{Collapsed rows: mean LCC size and mean edge count of the generated graphs against the test graphs (test: $\bar n$ / $\bar E$).}
\label{tab:collapsed}
\begin{tabular}{llrrl}
\toprule
Dataset (test $\bar n$ / $\bar E$) & Model & LCC $\bar n$ & $\bar E$ & Reason \\
\midrule
MMSBM (249 / 9{,}144) & GraphVAE & 1 & 0 & empty \\
 & LGD & 1 & 0 & empty \\
LFR (350 / 2{,}044) & GraphVAE & 1 & 0 & empty \\
 & SPECTRE & 42 & 191 & fragmented \\
 & VGAE & 346 & 31{,}679 & $15\times$ edges \\
 & EDP-GNN & 360 & 43{,}119 & $21\times$ edges \\
 & LGD & 223 & 28{,}092 & $14\times$ edges \\
CiteSeer-Ego (145 / 328) & GraphVAE & 1 & 0 & empty \\
 & VGAE & 148 & 7{,}213 & $22\times$ edges \\
 & EDP-GNN & 178 & 20{,}888 & $64\times$ edges \\
Protein (258 / 632) & LGD & 254 & 37{,}573 & $59\times$ edges \\
 & SPECTRE & 4 & 4 & fragmented$^{a}$ \\
\bottomrule
\end{tabular}
\\[2pt]\footnotesize $^{a}$ Two fresh draws of the reported checkpoint; the reported draw has LCC $\bar n = 203$ and is equally separable ($\mathrm{C2ST}_\Delta = 0.500$).
\end{table}

\paragraph{DiGress.} Running the official DiGress code on MMSBM and LFR initially exhausted host RAM (234\,GiB in preprocessing) due to an implementation bug: the dataset-information object keeps a reference to the whole data module, every graph of every split, and PyTorch Lightning copies it into the saved hyperparameters before the first training step. Our "DiGress-lean" version drops this reference once it has been used and keeps the extra-feature buffers out of the saved hyperparameters; the model, loss, noise schedule and sampler are unchanged, and a seeded CPU check gives identical losses and weights with and without it. With it, Protein needs 0.6\,GB of host memory after setup and 9.6\,GB of GPU memory per 499-node graph at batch size 1. Every DiGress row uses DiGress-lean except CiteSeer-Ego, whose run used the original entry point, which fits on that dataset. DiGress's cycle-count features are differences of traces of powers of the adjacency matrix; on dense graphs they exceed the exact-integer range of float32 (about $10^8$ per node on MMSBM), so small counts came out negative and stopped training at DiGress's own check. An opt-in switch computes them in float64, used on MMSBM and LFR.

\subsection{Sampling protocol}
\label{app:sampling}
Sampled-graph counts equal the test-set size for each dataset. Node counts $n$ are drawn from the empirical training-set distribution (exceptions above). All scoring is on the largest connected component (LCC) of each graph. Every model is timed on one NVIDIA RTX 6000 Ada from the checkpoint of its table row, at batch size 1, at batch size 8 and at the largest batch up to 256 that fits in GPU memory; the tables report the fastest, in seconds per graph (s/G). Times cover the whole generative pipeline at the settings of the reported samples (all denoising or Langevin steps, decoding and edge sampling), with $n$ drawn from the training sizes, and exclude model loading. For EDP-GNN and LGD each batch size is timed on one batch after a warm-up batch. Table~\ref{tab:sampling-bs} gives the batch size of each reported s/G.

\begin{table}[h]
\centering
\small
\caption{Batch size of the fastest timing (the s/G reported in the result tables).}
\label{tab:sampling-bs}
\begin{tabular}{lcccc}
\toprule
Model & MMSBM & LFR & CiteSeer-Ego & Protein \\
\midrule
VGAE & 64 & 8 & 1 & 1 \\
GraphVAE & 128 & 128 & 64 & 128 \\
SPECTRE & 8 & 8 & 8 & 8 \\
IGNR & 8 & 8 & 8 & 8 \\
GraphRNN & 256 & 256 & 256 & 256 \\
GRAN & 64 & 64 & 64 & 64 \\
EDP-GNN & 8 & 8 & 4 & 8 \\
LGD & 8 & 16 & 8 & 8 \\
DiGress & 8 & 8 & 8 & 1 \\
SparseDiff & 32 & 16 & 8 & 8 \\
EDGE & 128 & 256 & 128 & 256 \\
ILE & \flagOOM & \flagOOM & 8 & 64 \\
xVGAE (with $z_0$ diffusion) & 256 & 256 & 256 & 256 \\
xVGAE without diffusion & 64 & 64 & 8 & 128 \\
\bottomrule
\end{tabular}

\end{table}

\subsection{Orbit metric on LFR and MMSBM}
\label{app:orbit-lfr}
The standard orbit kernel compares two graphs by a Gaussian kernel of width $30$ on the TV distance between their node-averaged ORCA count vectors. Per-node counts grow like the cube of the degree, so these vectors are far apart: between two MMSBM test graphs the median distance is about $2\times10^5$ and the kernel is numerically $0$. When every off-diagonal kernel value is $0$, the biased $\mathrm{MMD}^2$ of any two sets of $m$ graphs equals $2/m$, whatever the graphs. We measure this with $1-\bar\kappa$, where $\bar\kappa$ is the mean kernel value between two distinct real graphs ($1$ = saturated): it reads $0.95$ on CiteSeer-Ego, $1.00$ on MMSBM and $1.00$ on LFR, and $0.24$ on Protein. On MMSBM the summed kernel returns exactly $2/m$ for held-out real graphs, for the same graphs after degree-preserving rewiring and for Erd\H{o}s--R\'enyi graphs with the same edge counts. Because a saturated orbit ratio is $2/m$ divided by $2/m$, it reads close to 1 for every model, including models whose graphs are far from the data. We therefore report the orbit column for comparability with prior work but flag it where it saturates (marked $^\dagger$), and rely on the degree, clustering, spectral and WL ratios and on $\mathrm{C2ST}$ there. 

\section{Hyperparameter Details}
\label{app:hyperparams}

\subsection{Configurations per Dataset}
\label{sec:hp_tables}

Table~\ref{tab:hp_sab} reports the autoencoder configurations for each xVGAE variant. All models share a 5-layer set-attention (SAB) encoder, hidden dimension $h = 64$, latent dimensions $d_z = 8$ and $d_{z_0} = 32$, $K_h = 8$ chunked bilinear heads (rank 4), a Gaussian mixture node prior conditioned on $z_0$, and a batch size of 8. Models are trained using AdamW with a learning rate of $5{\times}10^{-4}$ and weight decay $10^{-5}$. Decoders use either 5 SAB layers or a pointwise MLP on $(z_0, z_i)$. 

Six of the seven diffusion-prior models follow our standard training schedule: $\beta_0 = 0.1$, a cosine learning rate schedule, and $\beta$ annealed from $10^{-4}$ to $1$ along a half-cosine after a warm-up phase (with $\beta_0$ scaled by $\beta$). On CiteSeer-Ego, xVGAE-SAB uses a fixed $\beta = \beta_0 = 1$ and a constant learning rate for 2000 epochs. The $z_0$ diffusion prior is subsequently fitted to the frozen encoder, with both autoencoder and diffusion checkpoints selected on validation (Appendix~\ref{app:compute}). For variants without a diffusion prior ($z_0 \sim \mathcal{N}(0, I)$), checkpoints are selected on validation: CiteSeer-Ego uses the xVGAE-MLP autoencoder, Protein and LFR use the xVGAE-SAB autoencoders, and MMSBM uses the dedicated configuration in the final row of Table~\ref{tab:hp_sab}.

\begin{table}[h]
\centering
\caption{%
  \textbf{xVGAE autoencoder configurations.} 
  All models use a 5-layer SAB encoder and either a 5-layer SAB or pointwise MLP decoder. Schedule triples denote warm-up / anneal / total epochs. $^{\ast}$Uses a 32-component Gaussian-mixture node prior, weight decay $10^{-8}$, and $z_0\sim\mathcal N(0,I)$.
}
\label{tab:hp_sab}
\resizebox{\textwidth}{!}{%
\begin{tabular}{llcccc}
\toprule
\textbf{Dataset} & \textbf{Row} & Encoder & Decoder & $\beta_0$ & $\beta$ schedule; LR schedule \\
\midrule
CiteSeer-Ego   & xVGAE-SAB & SAB, 5 & SAB, 5 & 1   & $\beta{=}1$, 2000 ep.; constant \\
CiteSeer-Ego   & xVGAE-MLP & SAB, 5 & MLP    & 0.1 & 16 / 46 / 1230; cosine  \\
MMSBM          & xVGAE-SAB & SAB, 5 & SAB, 5 & 0.1 & 12 / 35 / 938; cosine  \\
MMSBM          & xVGAE-MLP & SAB, 5 & MLP    & 0.1 & 12 / 35 / 938; cosine  \\
LFR            & xVGAE-SAB & SAB, 5 & SAB, 5 & 0.1 & 12 / 35 / 938; cosine \\
LFR            & xVGAE-MLP & SAB, 5 & MLP    & 0.1 & 12 / 35 / 938; cosine \\
Protein (D\&D) & xVGAE-SAB & SAB, 5 & SAB, 5 & 0.1 & 13 / 38 / 1014; cosine \\
\midrule
MMSBM          & without diffusion$^{\ast}$ & SAB, 5 & MLP & 1 & $\beta$ fixed, 5000 ep.; constant \\
\bottomrule
\end{tabular}%
}
\end{table}

\subsection{\texorpdfstring{Chunked Bilinear Decoder and $W$ Parameterisations}{Chunked Bilinear Decoder and W Parameterisations}}
\label{sec:bilinear_decoder}

The logit of Eq.~\eqref{eq:graphon-head} is a mixture of $K_h$ bilinear forms, each acting on its own slice of the node embedding $h_i \in \R^{d_h}$:
\begin{equation}
  s_{ij} \;=\; \sum_{k=1}^{K_h} \alpha_k(z_0)\, h_i^{(k)\top} W_k(z_0)\, h_j^{(k)} \;+\; b_0(z_0) + c(h_i) + c(h_j),
  \qquad \omega_\theta(z_0, h_i, h_j) = \sigma(s_{ij}),
  \label{eq:bilinear_logit}
\end{equation}
where $h_i^{(k)} \in \R^{d_h/K_h}$ is the $k$-th of $K_h$ disjoint chunks of $h_i$ (for example, $d_h = 64$, $K_h = 8$: eight dimensions each), $\alpha(z_0)$ are softmax-normalised head weights, $b_0(z_0)$ is a per-graph bias and $c$ is a per-node bias shared by all nodes. The per-node bias and a symmetric pair term are used by the MLP decoder only.

\paragraph{Chunking.} The chunks are what make the heads distinct: applied to the full embedding, $\sum_k \alpha_k\, h_i^\top W_k h_j = h_i^\top \bigl(\sum_k \alpha_k W_k\bigr) h_j$ is a single bilinear form.

\paragraph{Interaction matrices.} Each head uses $W_k(z_0) = W_k^{\mathrm{base}} + W_k^{\mathrm{mod}}(z_0)$, where $W_k^{\mathrm{base}}$ is a symmetric matrix shared across the dataset (normalised by its spectral norm, detached from the gradient) and $W_k^{\mathrm{mod}}$ is predicted from $z_0$, with $d_k = d_h/K_h$. Four forms of $W_k^{\mathrm{mod}}$ are available, nested as follows: \texttt{full}, a general symmetric $d_k \times d_k$ matrix; \texttt{low-rank}, $P_a P_a^\top - P_r P_r^\top$ with $P_a, P_r \in \R^{d_k \times r}$, which equals \texttt{full} when $r = d_k$ (every symmetric matrix is a difference of two positive semi-definite ones) and allows both assortative and disassortative patterns; \texttt{diag}, the same difference with diagonal factors in learned bases, a cheaper special case of \texttt{low-rank}; and \texttt{psd}, $LL^\top$, the special case $P_r = 0$, which allows assortative patterns only. All reported models use \texttt{low-rank} with $r = 4$.

\section{Latent spaces}
\label{app:latent}
\subsection{Per-graph latent variables}

\begin{figure}[h!]
    \centering
    \includegraphics[width=1\linewidth]{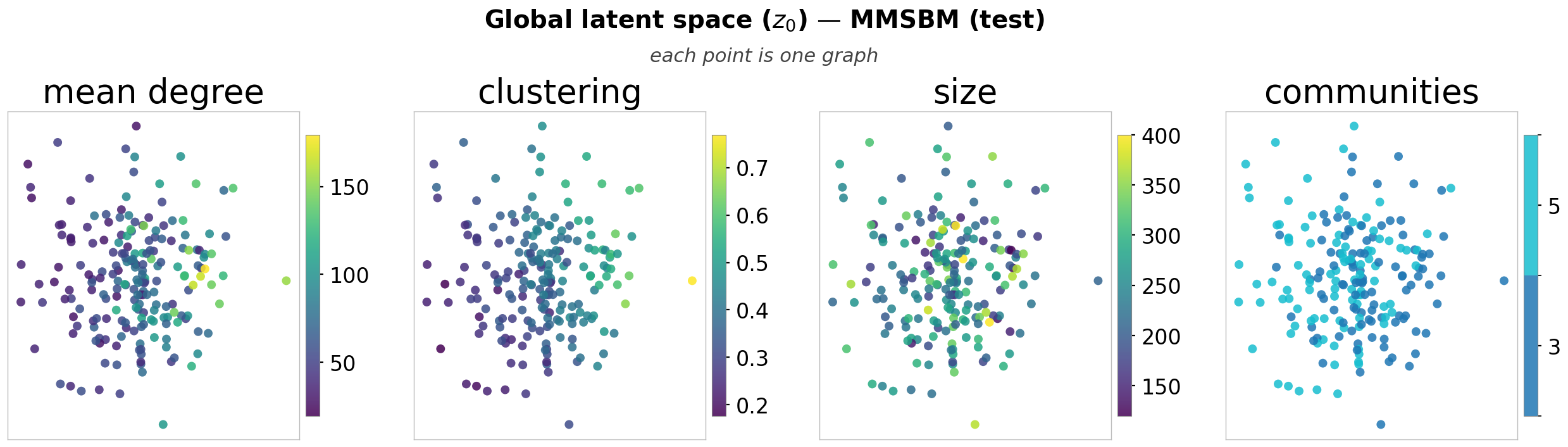}
    \caption{The latent space of the global variable $z_0$ embeds structural information of MMSBM graphs. The plots show a clear association between the global variables and the graphs' average degree, clustering coefficient, size and number of communities.}
    \label{fig:mmsbm_global}
\end{figure}

\begin{figure}[h!]
    \centering
    \includegraphics[width=1\linewidth]{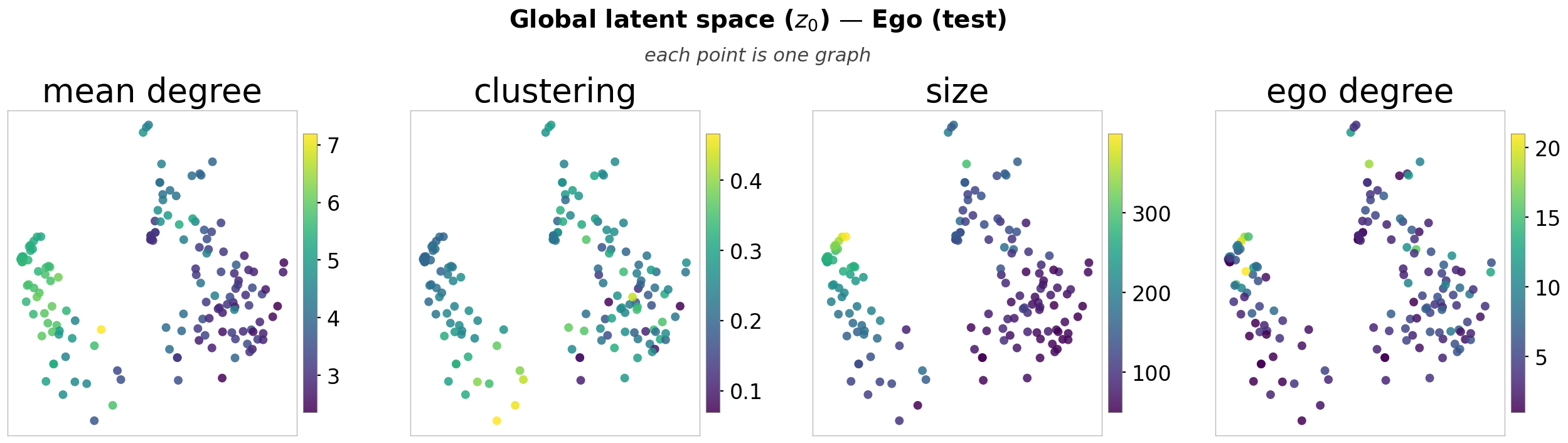}
    \caption{The latent space of the global variable $z_0$ embeds structural information of Ego graphs. The plots show a clear association between the global variables and the graphs' average degree, clustering coefficient, size and the degree of the Ego node.}
    \label{fig:ego_global}
\end{figure}

\begin{figure}[h!]
    \centering
    \includegraphics[width=1\linewidth]{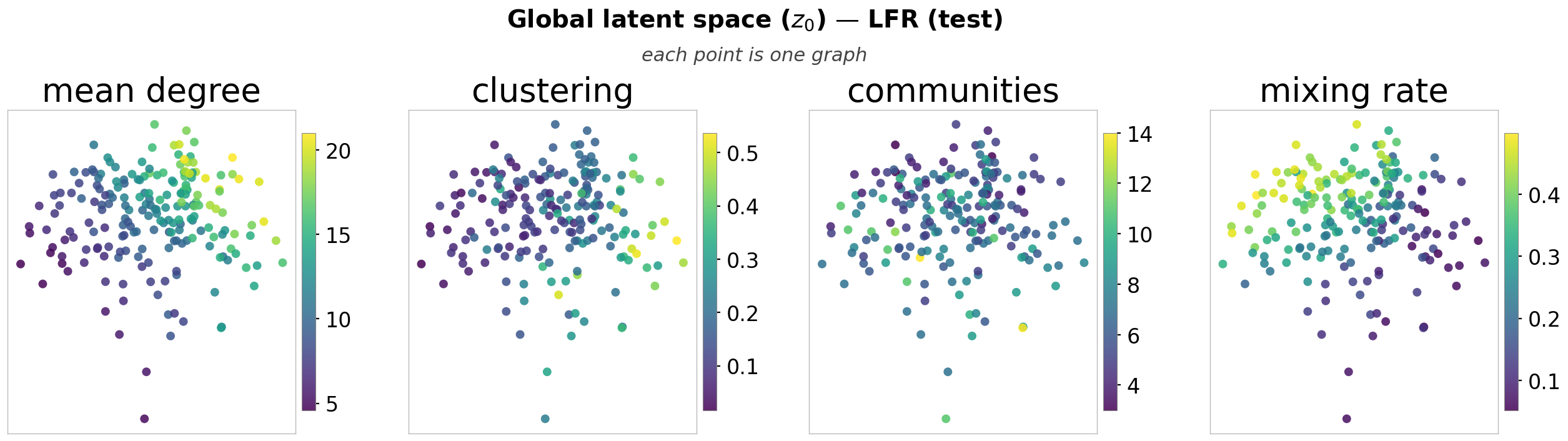}
    \caption{The latent space of the global variable $z_0$ embeds structural information of LFR graphs. The plots show a clear association between the global variables and the graphs' average degree, clustering coefficient, number of communities and mixing parameter.}
    \label{fig:lfr_global}
\end{figure}

\begin{figure}[h!]
    \centering
    \includegraphics[width=1\linewidth]{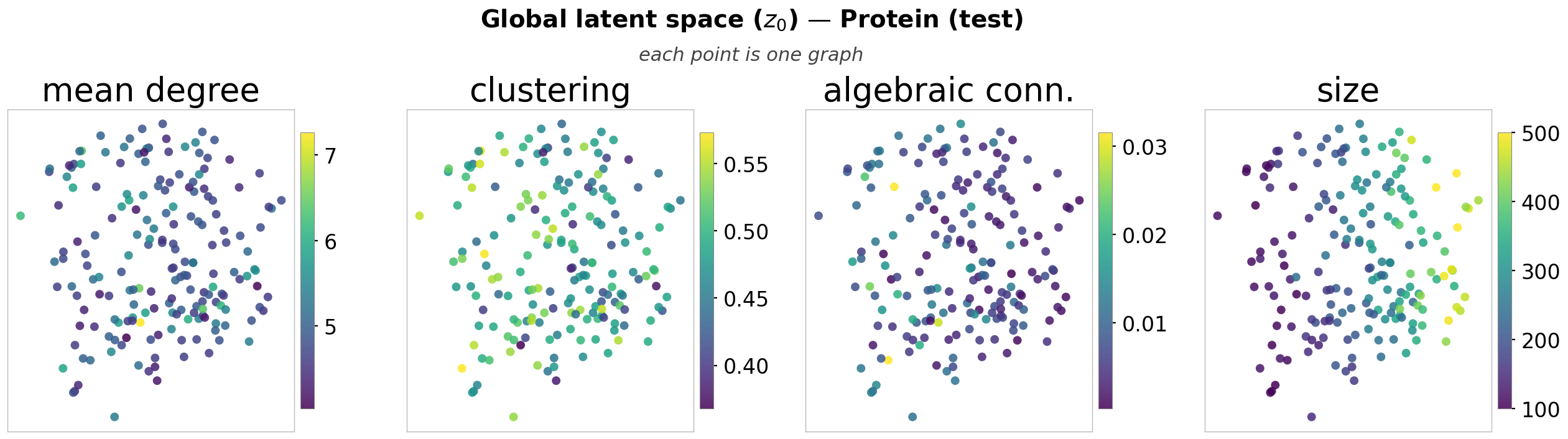}
    \caption{The latent space of the global variable $z_0$ embeds structural information of Protein graphs. The plots show a clear association between the global variables and the graphs' average degree, clustering coefficient, algebraic connectivity and graph size.}
    \label{fig:pro_global}
\end{figure}
\clearpage

\subsection{Per-node latent variables}

\begin{figure}[h!]
    \centering
    \includegraphics[width=1\linewidth]{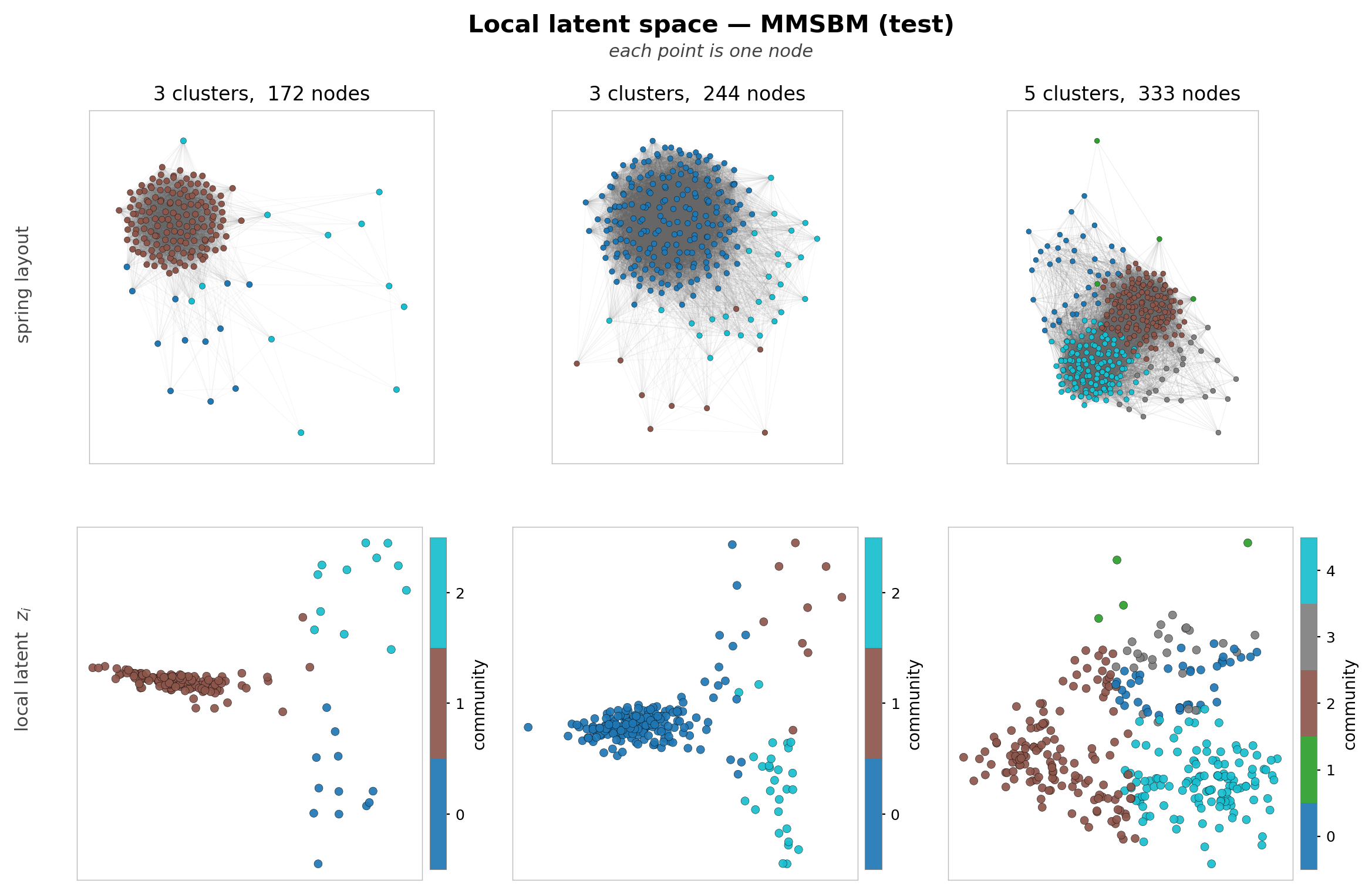}
    \caption{\textit{Top:} Three graphs from the test set of MMSBM (spring layout). The nodes are colored by their strongest community membership. \textit{Bottom:} PCA 2D projection of the latent space for the encoded local variables, demonstrating that the latent space of our xVGAE clusters together nodes with similar membership distribution even though it was only exposed to the adjacency matrix.}
    \label{fig:mmsbm_local}
\end{figure}

\begin{figure}[h!]
    \centering
    \includegraphics[width=1\linewidth]{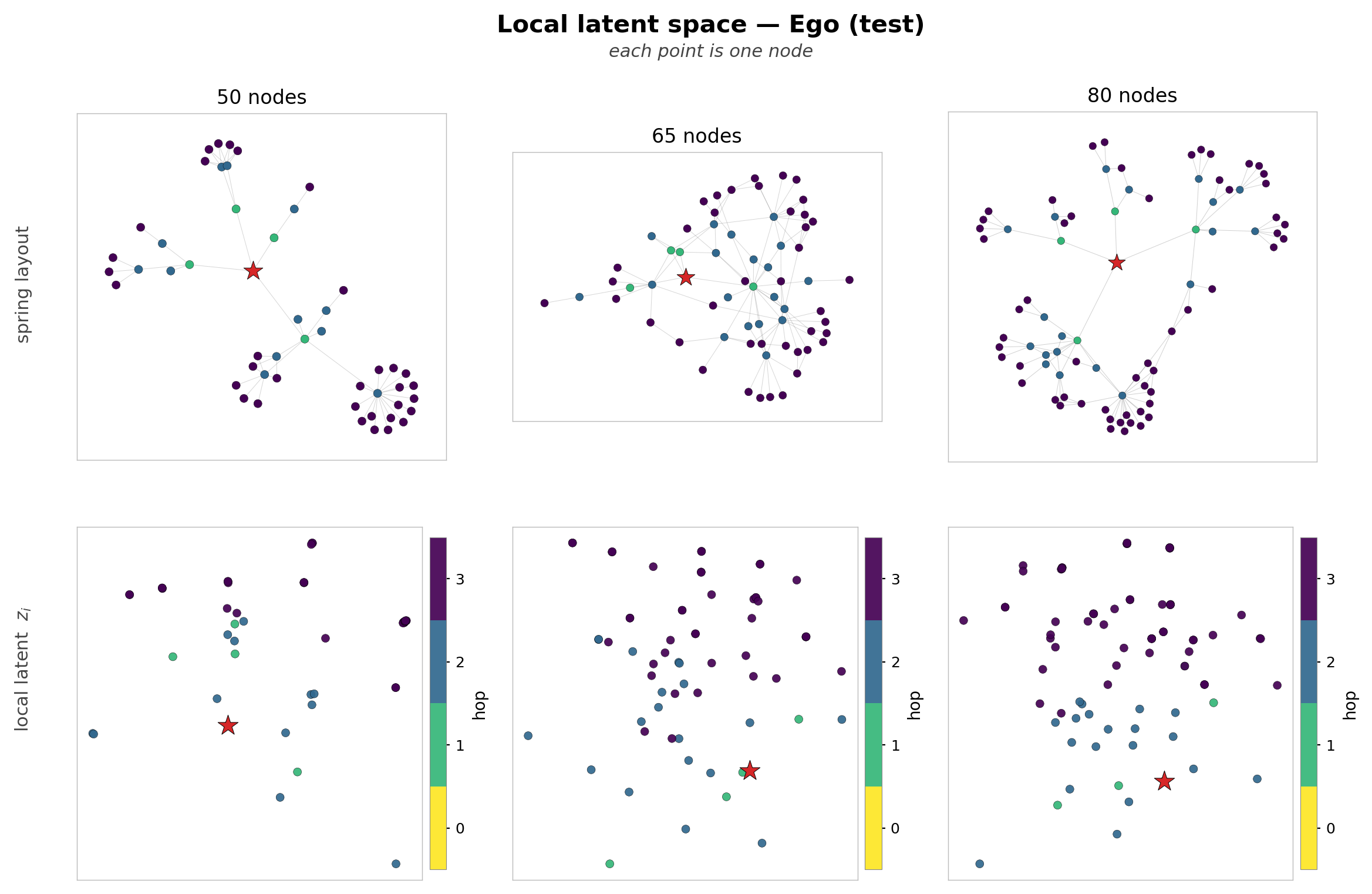}
    \caption{\textit{Top:} Three graphs from the test set of Ego (spring layout). The nodes are colored based on the number of hops separating them from the Ego node, represented with a star. \textit{Bottom:} PCA 2D projection of the latent space for the encoded local variables, demonstrating that the latent space of our xVGAE maps nodes based on their distance from the Ego node, even though it does not know which node is the Ego.}
    \label{fig:ego_local}
\end{figure}

\begin{figure}[h!]
    \centering
    \includegraphics[width=1\linewidth]{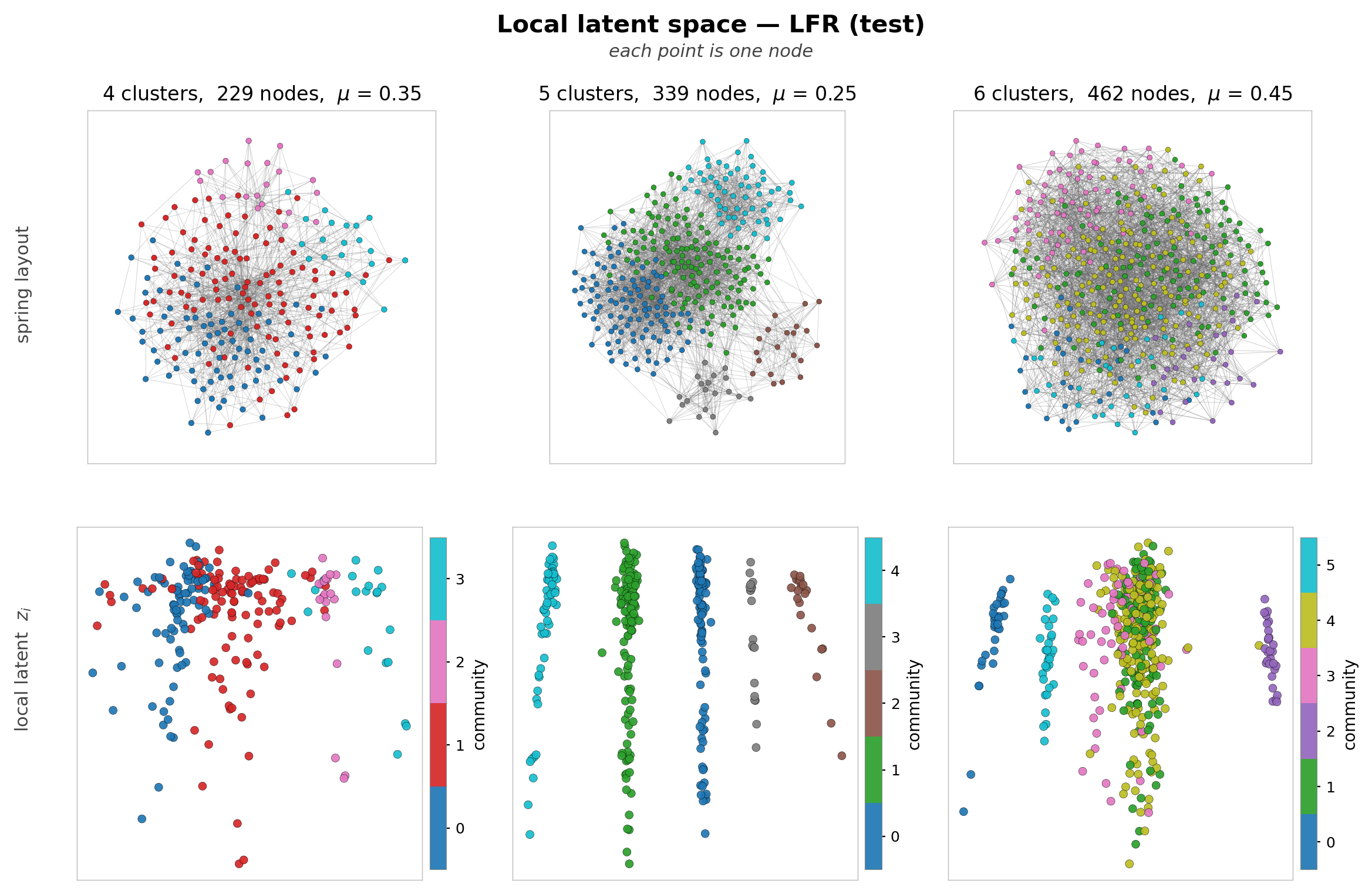}
    \caption{\textit{Top:} Three graphs from the test set of LFR (spring layout). The nodes are colored by their community membership, showing that the latent space of our xVGAE clusters together nodes with the same community membership and orders them by relative degree, even though it was only exposed to the adjacency matrix.
 \textit{Bottom:} PCA 2D projection of the latent space for the encoded local variables, demonstrating that the latent space of our xVGAE clusters together nodes with the same community membership even though it was only exposed to the adjacency matrix.}
    \label{fig:lfr_local}    
\end{figure}

\begin{figure}[h!]
    \centering
    \includegraphics[width=1\linewidth]{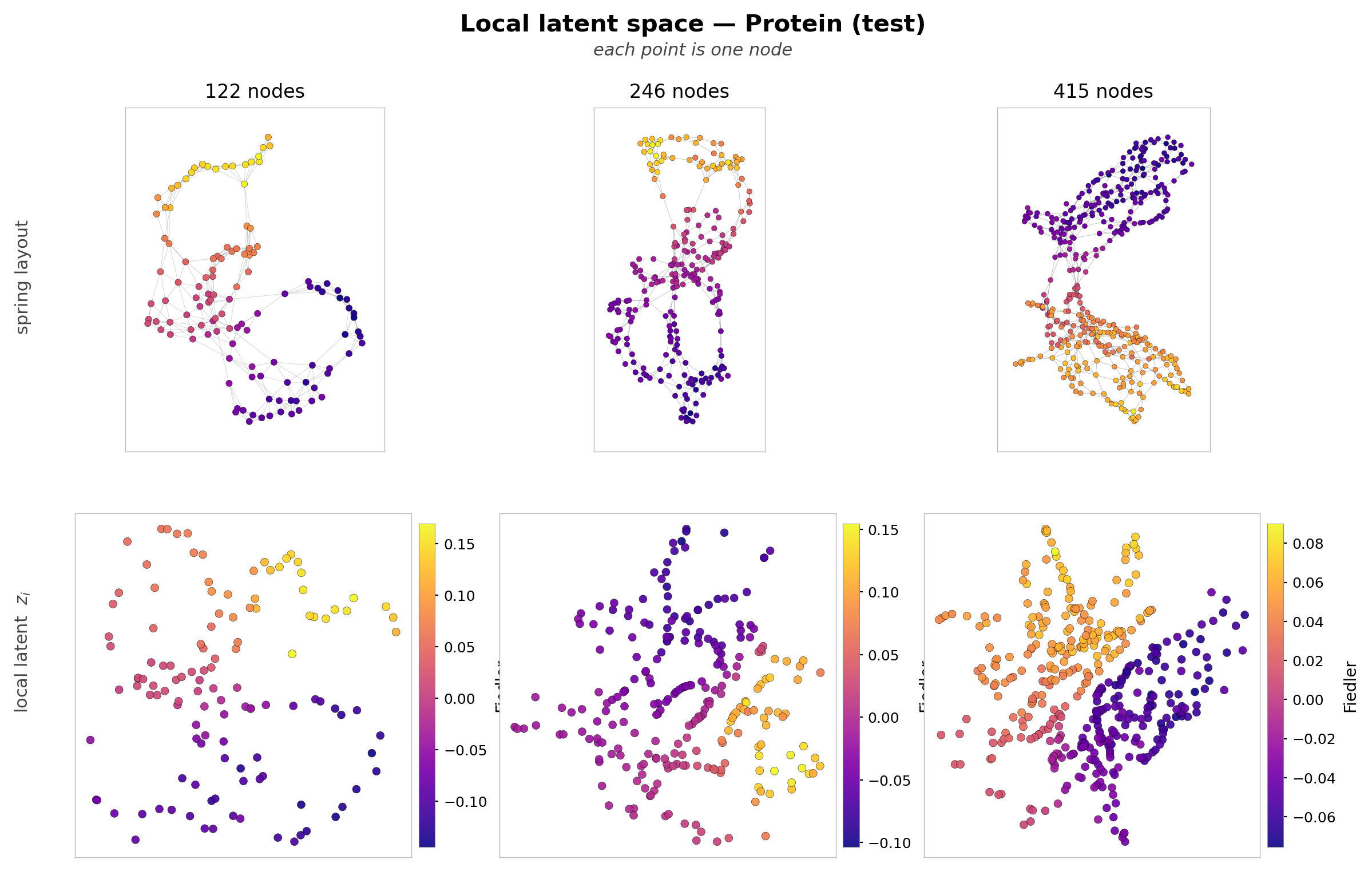}
    \caption{\textit{Top:} Three graphs from the test set of Protein (spring layout). The nodes are colored by Fiedler vector. \textit{Bottom:} PCA 2D projection of the latent space for the encoded local variables, demonstrating that the latent space of our xVGAE contains meaningful information given the clustering colors based on the Fiedler vector.}
    \label{fig:pro_local}    
\end{figure}

\clearpage

\section{Illustrated Graphs with spring layout}
\label{app:topologies}
\begin{figure}[h!]
    \centering
    \includegraphics[width=0.65\linewidth]{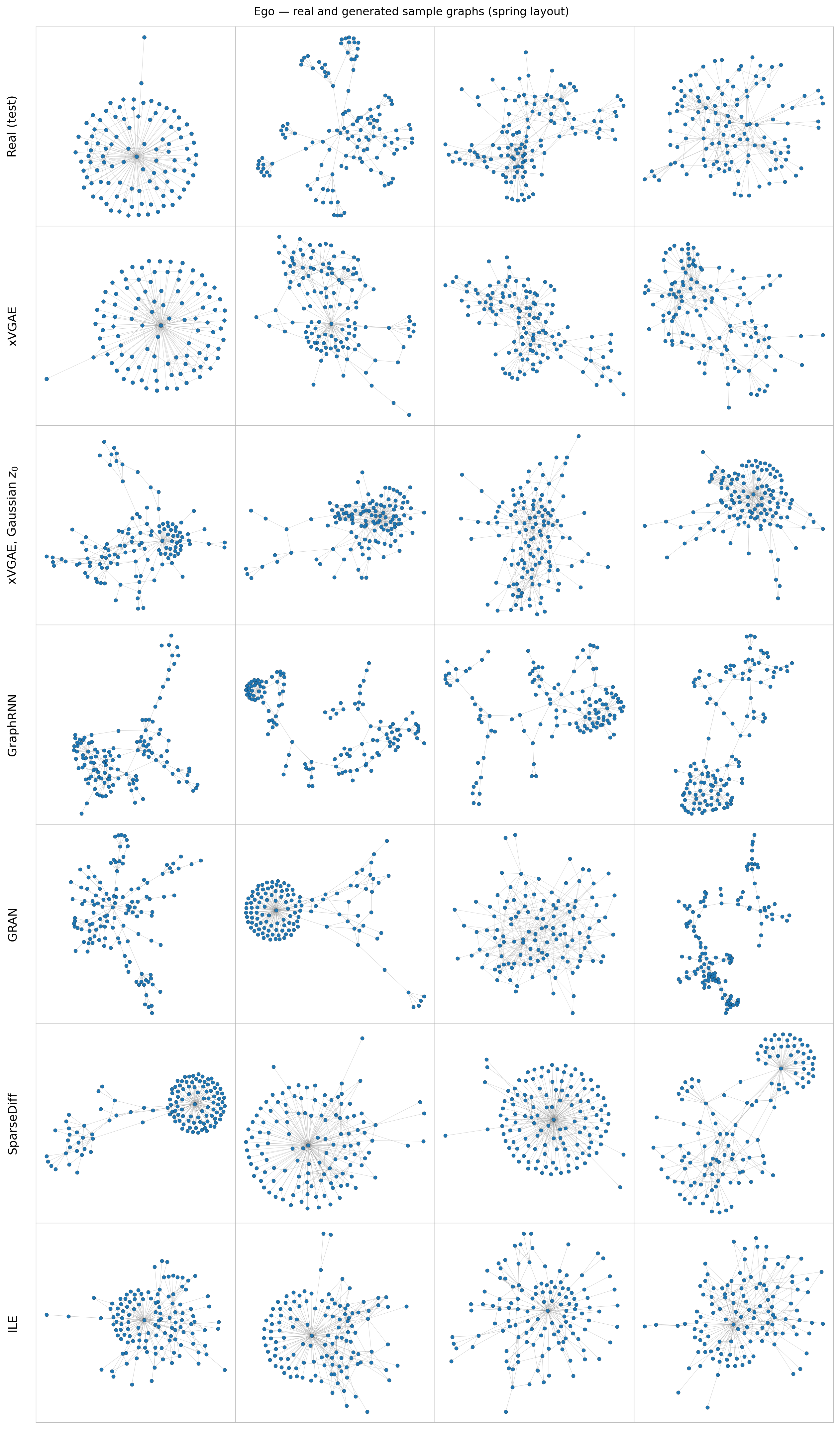}
    \caption{Examples of generated Ego graphs (spring layout) from our model, main baselines and version of xVGAE without global diffusion.}
    \label{fig:ego_topologies}
\end{figure}

\begin{figure}
    \centering
    \includegraphics[width=0.7\linewidth]{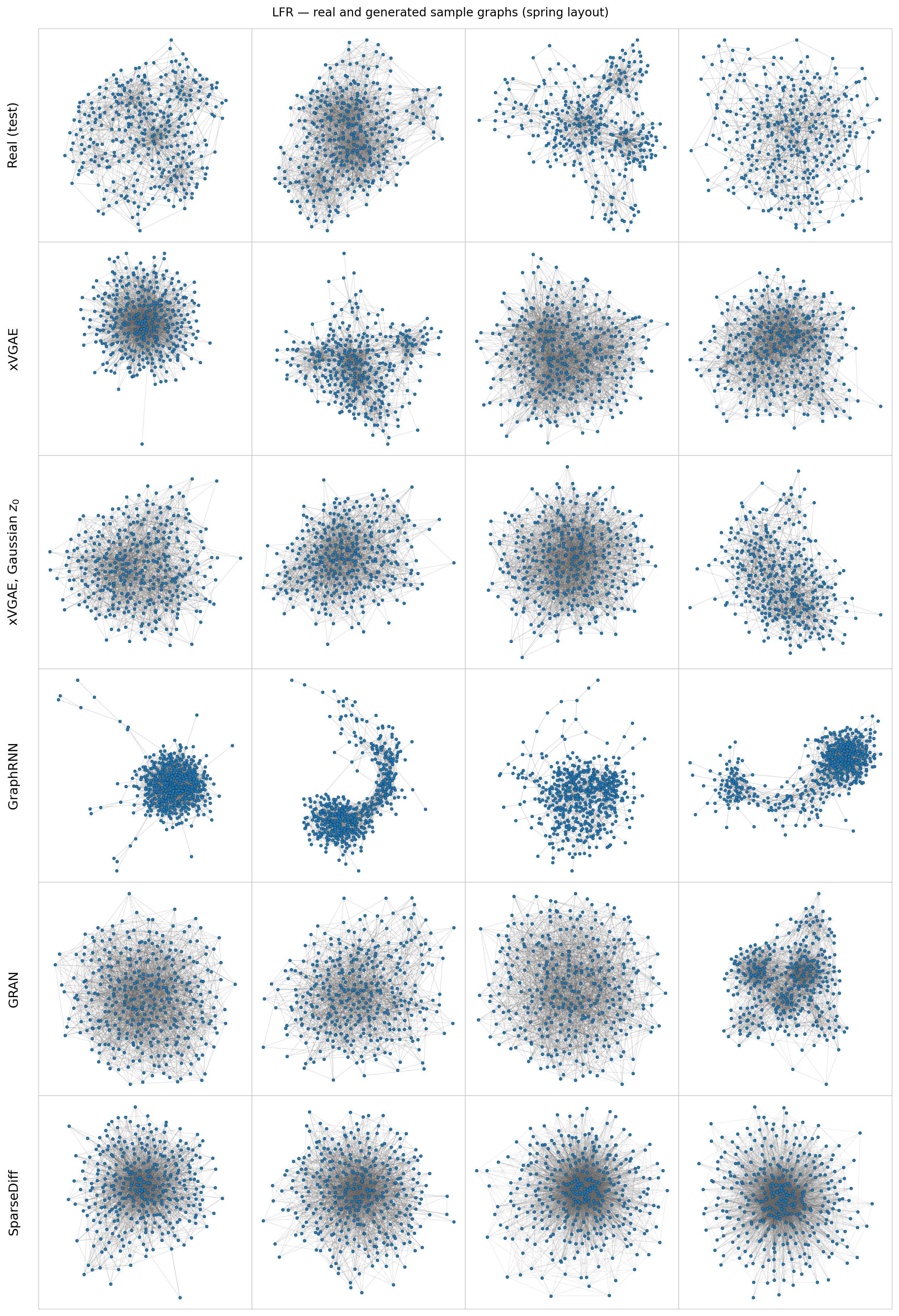}
    \caption{Examples of generated LFR graphs (spring layout) from our model, main baselines and version of xVGAE without global diffusion.}
    \label{fig:lfr_topologies}
\end{figure}

\begin{figure}
    \centering
    \includegraphics[width=0.7\linewidth]{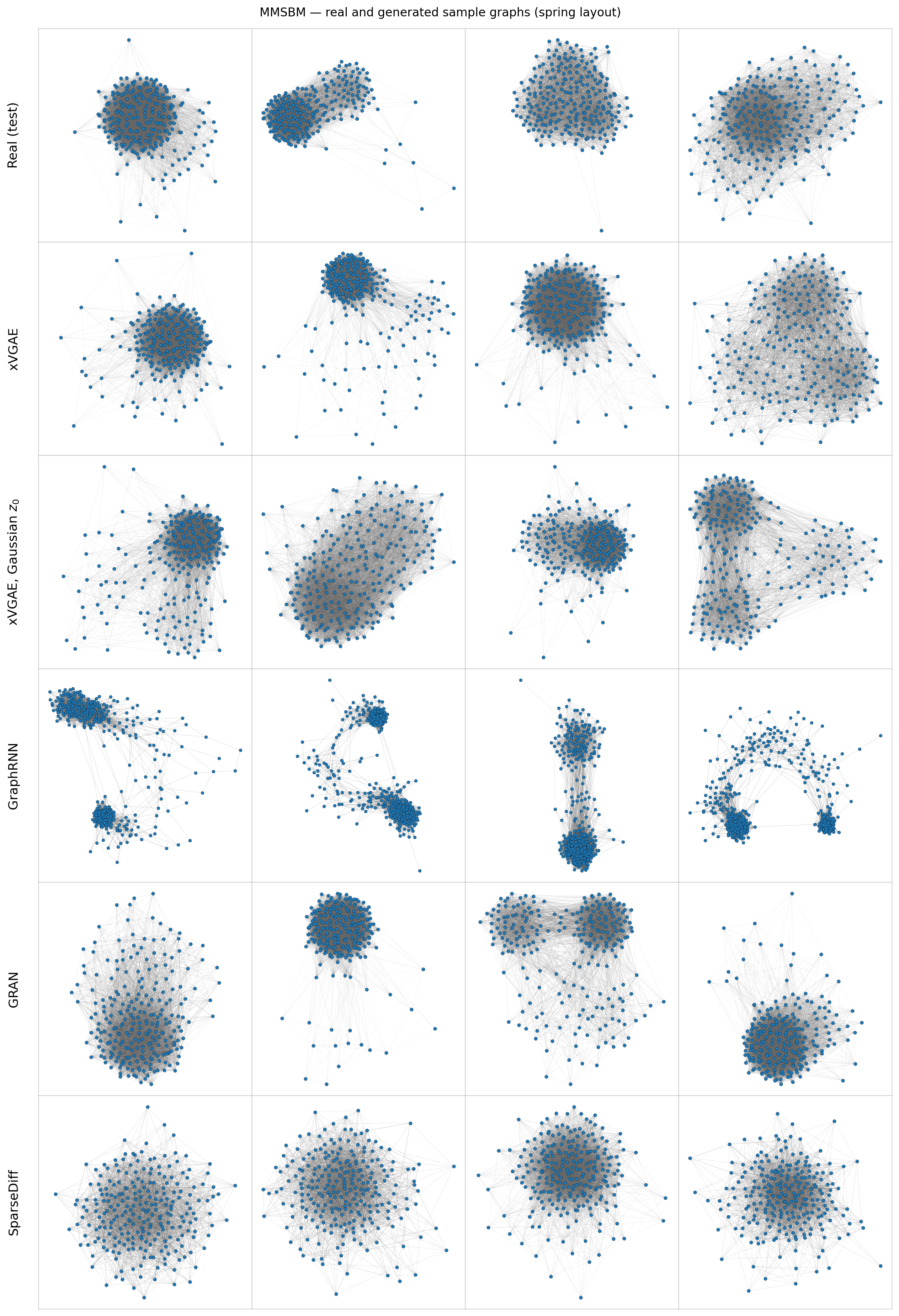}
    \caption{Examples of generated MMSBM graphs (spring layout) from our model, main baselines and version of xVGAE without global diffusion.}
    \label{fig:mmsbm_topologies}
\end{figure}

\begin{figure}
    \centering
    \includegraphics[width=0.7\linewidth]{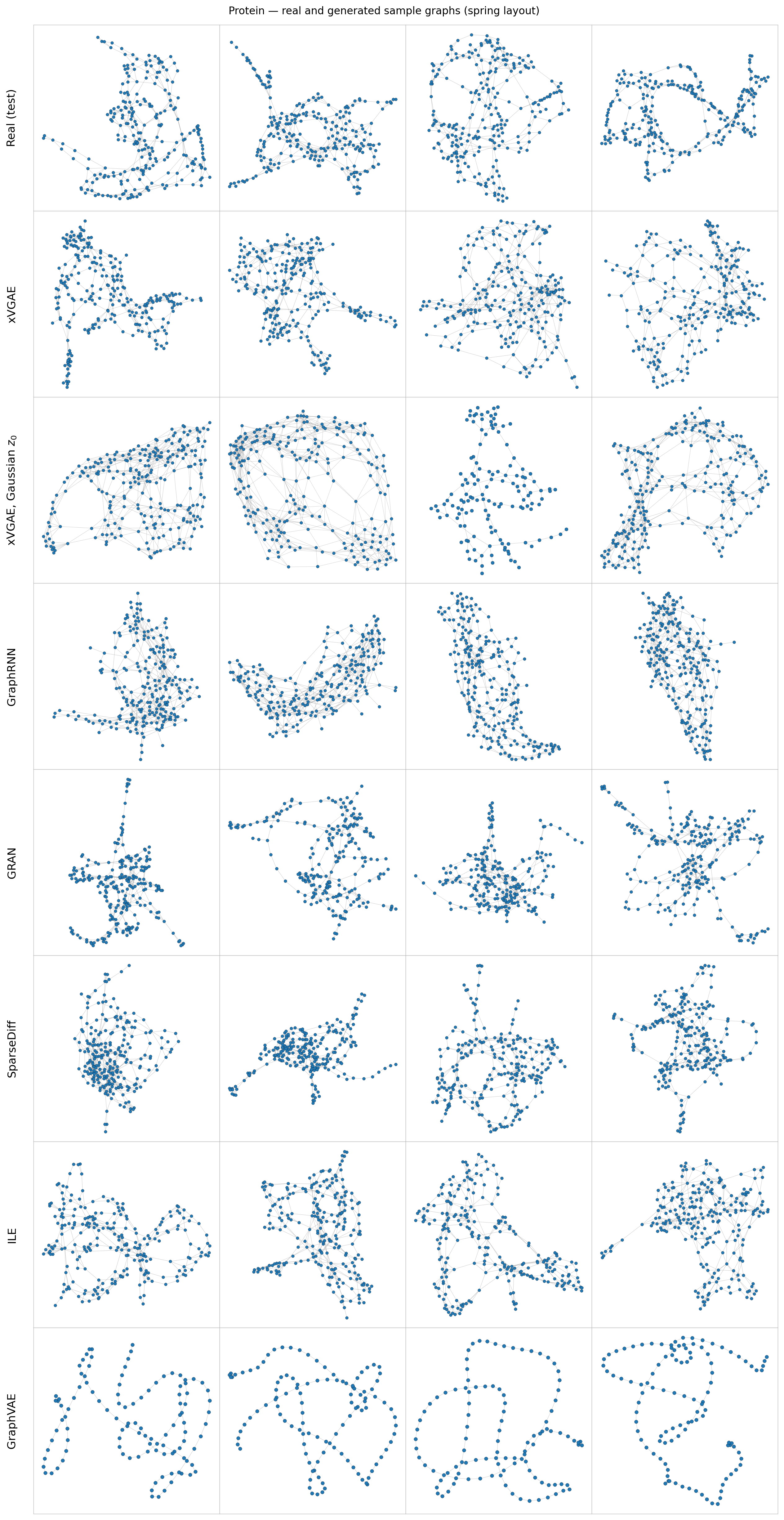}
    \caption{Examples of generated Protein graphs (spring layout) from our model, main baselines and version of xVGAE without global diffusion.}
    \label{fig:protein_topologies}
\end{figure}
\appendix



\end{document}